\documentclass[a4paper,fleqn]{cas-dc}

\usepackage[numbers]{natbib}

\newcommand{\WRP}{\par\qquad\qquad\qquad\(\hookrightarrow\)\enspace}
\usepackage[mathscr]{eucal}
\usepackage{amsmath,mathtools}
\usepackage{amssymb}
\usepackage{lipsum}
\usepackage{algorithmicx}
\usepackage[ruled]{algorithm}
\usepackage[noend]{algpseudocode}
\usepackage{subfig}

\usepackage[switch]{lineno} 
\modulolinenumbers[2]

\newcommand*{\img}[1]{%
    \raisebox{0\baselineskip}{%
        \includegraphics[
        height=0.6\baselineskip,
        width=0.6\baselineskip,
        keepaspectratio,
        ]{#1}%
    }%
}

\def\tsc#1{\csdef{#1}{\textsc{\lowercase{#1}}\xspace}}
\tsc{WGM}
\tsc{QE}
\tsc{EP}
\tsc{PMS}
\tsc{BEC}
\tsc{DE}
\begin{document}
\let\WriteBookmarks\relax
\def\floatpagepagefraction{1}
\def\textpagefraction{.001}
% \shorttitle{Model-agnostic interpretation by visualization of feature perturbations}
% \shortauthors{Marcílio-Jr et al.}
\shorttitle{}
\shortauthors{}

\title [mode = title]{Model-agnostic interpretation by visualization of feature perturbations}                      
%\tnotemark[1,2]

%\tnotetext[1]{This document is the results of the research
%   project funded by the National Science Foundation.}

%\tnotetext[2]{The second title footnote which is a longer text matter
%   to fill through the whole text width and overflow into
%   another line in the footnotes area of the first page.}

\author[1]{Wilson E. Marcílio-Jr}
\author[1]{Danilo M. Eler}
\author[2]{Fabrício Breve}
%\cormark[1]
%\fnmark[1]
%\ead{cvr_1@tug.org.in}
%\ead[url]{www.cvr.cc, cvr@sayahna.org}

%\credit{Conceptualization of this study, Methodology, Software}

\address[1]{Faculty of Sciences and Technology, São Paulo State University (UNESP), Presidente Prudente, SP 19060-900, Brazil}
\address[2]{Institute of Geosciences and Exact Sciences, São Paulo State University (UNESP), Rio Claro, SP 13506-900, Brazil}

%\author[2,4]{Han Theh Thanh}[style=chinese]

%\author[2,3]{CV Rajagopal}[%
 %  role=Co-ordinator,
%   suffix=Jr,
 %  ]
%\fnmark[2]
%\ead{cvr3@sayahna.org}
%\ead[URL]{www.sayahna.org}

%\credit{Data curation, Writing - Original draft preparation}

%\address[2]{Sayahna Foundation, Jagathy, Trivandrum 695014, India}

%\author%
%[1,3]
%{Rishi T.}
%\cormark[2]
%\fnmark[1,3]
\cortext[cor1]{E-mails: wilson.marcilio@unesp.br, danilo.eler@unesp.br, fabricio.breve@unesp.br} 

\begin{abstract}
Interpretation of machine learning models has become one of the most important research topics due to the necessity of maintaining control and avoiding bias in these algorithms. Since many machine learning algorithms are published every day, there is a need for novel model-agnostic interpretation approaches that could be used to interpret a great variety of algorithms. Thus, one advantageous way to interpret machine learning models is to feed different input data to understand the changes in the prediction. Using such an approach, practitioners can define relations among data patterns and a model's decision. This work proposes a model-agnostic interpretation approach that uses visualization of feature perturbations induced by the PSO algorithm. We validate our approach on publicly available datasets, showing the capability to enhance the interpretation of different classifiers while yielding very stable results compared with state-of-the-art algorithms.
\end{abstract}

% \begin{graphicalabstract}
% \includegraphics[width=\linewidth]{figs/visualization-design/tool.pdf}
% \end{graphicalabstract}

% \begin{highlights}
% \item A new approach for machine learning model's interpretation is presented
% \item Particle swarm optimization helps at interpret model decisions
% \item Novel visualization methods are presented to analyze classifier confusion
% \end{highlights}


\maketitle

% \linenumbers

\section{Introduction}

Machine learning (ML) algorithms have been achieving unprecedented capability in various tasks, such as in Natural Language Processing~\citep{Peters2018, Devlin2018}, Computer Vision~\citep{Szegedy2015, Krizhevsky2016, He2016, Chollet2017}, and others. Nevertheless, the employment of ML models on applications where the consequences of mistakes could be catastrophic for organizations~\citep{Luo2016} has required the necessity to include model interpretation capabilities when developing ML solutions. Interpretability increases the performance of human-model teams~\citep{Bansal2019} and improves the ability of practitioners to debug models~\citep{Kulesza2015}, leading to better hyper-parameter tuning. 

To this end, Explainable Artificial Intelligence (XAI) efforts focus on the interpretability side-by-side of model performance. In this case, the literature presents various approaches to decrease the lack of interpretation faced by complex models. For example, the surrogate models mimic machine learning models to provide global or local explanations~\citep{DoshiVelez2017, DoshiVelez2018}. At the same time, visualization techniques usually focus on specific models to devise graphical representations and enhance understanding of models' decisions and functionality~\citep{Krause2016, Pezzotti2018, MarcilioJr2020}. Such visualization techniques can emphasize nuances and provide insights into the execution of these black-box models to indicate a bias towards a specific class. Another class of XAI methods that has been receiving much attention is the model-agnostic explanation methods, which create surrogate models to return the contribution of each feature to a model's prediction~\citep{Lundberg2017, Ribeiro2016}. Using these strategies, specialists could look at the model's prediction, analyze which features contribute the most, and further assess if these contributions make sense with the problem domain. For example, a doctor would look to the features considered after a model predicts a tumor as benign or not. Then, such prediction could be assessed as reliable or not even if the model's prediction was correct. The characteristic of gaining insights even if a model does not have high performance is one of the most important and valuable aspects of XAI methods. Although much research focuses on interpretability aspects of machine learning models, only a few provide model-agnostic interpretation based on visual exploration~\citep{Zhang2019, Hinterreiter2020}. These existing approaches still lack flexibility regarding whether they could be applied to interpret classification, regression tasks~\citep{Hinterreiter2020}, or assign feature importance that could not correspond to the true importance of machine learning models~\citep{Zhang2019}.

In this work, we propose a novel model-agnostic interpretation approach based on the visualization of feature perturbations generated by the PSO algorithm~\citep{Olsson2010}. Our approach uses the PSO algorithm to minimize a function that induces the most change in a model's prediction. Then, with carefully chosen visual encodings, the decisions of a model can be inspected. We use a prototype tool with coordinated views to visualize the similarity between classes using a hybrid visualization of a radial layout and node-link diagram, besides using detailed and summary visualizations based on dot plot to encode the importance of features to a model's prediction. Our tool focuses on visualizing the PSO particles representing weights and the model's performance after multiplying these weights using the x-axis and model performance using a color scale. We also provide histograms of feature values to help users compare the feature importance and their actual distributions. Finally, we design a summary visualization to encode the PSO execution using a strategy based on heatmaps and a node-link strategy that emphasizes the relationship among dataset classes according to how PSO particles (perturbing weights) affect their correct classification. Our methodology is validated through several case studies by showing its applicability to understanding machine learning models. Then, we numerically evaluate our approach according to feature importance, showing that our method yields very stable results across different datasets and surpass a few well-established methods in the sense of assigning correct importance to features. This is the first research study exploring the PSO algorithm to generate perturbation of feature spaces to interpret machine learning models to the best of our knowledge. Moreover, this is the first research study visualizing the execution of the PSO algorithm.

\begin{itemize}
    \item Support machine learning models and feature spaces interpretation using the PSO algorithm.;
    \item Novel visual metaphors to visualize PSO execution and to interpret machine learning models.
\end{itemize}

This work is organized as follows: Section~\ref{sec:related-works} presents the related works on model explainability; in Section~\ref{sec:methodology} we delineate our methodology; in Section~\ref{sec:visualization-design}, we explain our visualization design; in Section~\ref{sec:use-cases} we show use cases to demonstrate how our method can be used to interpret a model's prediction; Section~\ref{sec:numerical_evaluation} provides a numerical evaluation of our technique regarding its ability to truly reflect the feature importance; discussions are provided in Section~\ref{sec:discussion}; we conclude our work in Section~\ref{sec:conclusion}.

\section{Related Works}
\label{sec:related-works}

The adoption of machine learning-based approaches in healthcare~\citep{Balagopalan2018, Esteva2017, Krause2016}, finance~\citep{Modarres2018}, government~\citep{Meijer2019}, and other areas must account for explainability factors. For example, doctors would like to understand a model's decisions after predicting a tumor as benign or not. Besides defining the interpretability of machine learning with scientific rigor~\citep{DoshiVelez2017, DoshiVelez2018} and identifying the human factors in model interpretability, such as practices, challenges, and needs~\citep{Hong2020}, the literature is focused on proposing novel techniques to interpret machine learning models.

These techniques are often categorized into two classes: global and local~\citep{DoshiVelez2018}. Global methods summarize the input features' contributions to the model as a whole~\citep{Lundberg2020}, which could deceive one understanding of the structures of the internal model's decisions. Such methods usually try to understand a model structure and functionality by applying input-output combinations to build a mental map of a model's decision, such as building surrogate models to measure the importance of features by adding perturbation. On the other hand, local interpretability methods explain each data observation separately. LIME~\citep{Ribeiro2016} and SHAP~\citep{Lundberg2017}, for example, propose model-agnostic explanations by explaining models' predictions through the contribution of features. In this case, this set of feature contributions help explain the model's prediction. Note that global interpretation for LIME and SHAP is achieved by averaging the local contributions. Other methods apply various model-agnostic approaches that require repeatedly executing the model for each explanation~\citep{Baehrens2010, Strumbelj2013, Datta2016}---note that LIME and SHAP also repeatedly execute the model.

Finally, the visualization community has been putting much effort into using graphical representations to help researchers interpret neural networks~\citep{Smilkov2017, Kahng2018} and understand deep learning models' training processes~\citep{Rauber2017, Pezzotti2018, MarcilioJr2020}, such as visualizing neurons' activations using heatmap representations~\citep{Clavien2019}.

In this work, we present a model-agnostic explanation approach in which perturbation weights are found through PSO. Our method searches for the minimum weight that induces the greater change in the model performance to find the importance of a feature. Besides, we present a visualization approach to interpreting the results generated by our algorithm.

\section{Methodology}
\label{sec:methodology}

This work aims to interpret a  model's prediction by understanding which decisions it took to classify data samples correctly or incorrectly. Such an understanding is achieved by inspecting the importance of the features regarding the classification outcome.

To devise an interpretation for a model's prediction, let us first discuss how this could be done using the confusion of a trained classifier (when a classifier assigns an incorrect class to a data sample). To explore the confusions of a classifier, we create perturbations on the test set (data samples used to evaluate a model's performance). These perturbations correspond to real numbers greater than zero multiplied by the feature values of the test set. Here, if a classifier maintains its performance after perturbing a specific feature, the classifier is stable to such a feature since perturbations do not influence the prediction. Thus, the feature perturbations can measure the stability of a model.

\begin{figure}[!htb]
 \centering % avoid the use of \begin{center}...\end{center} and use \centering instead (more compact)
 \includegraphics[width=\columnwidth]{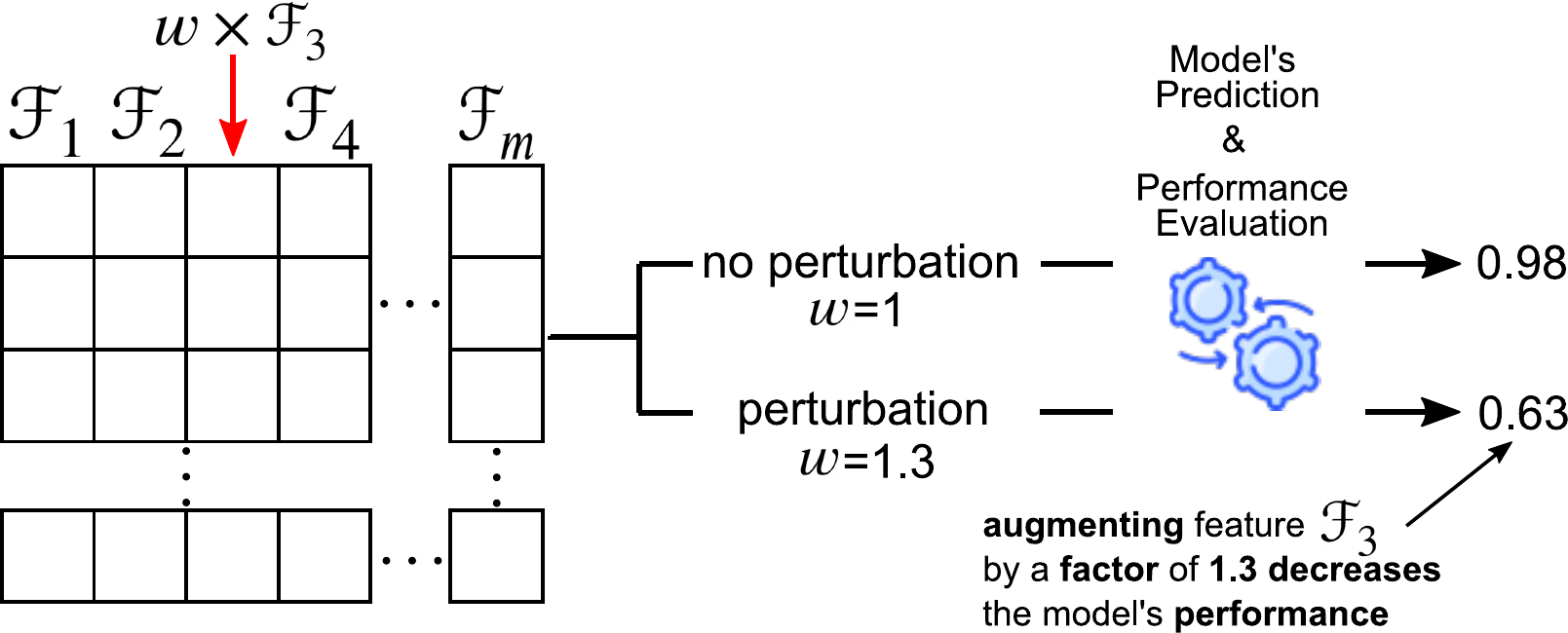}
 \caption{Perturbing a feature with weight. After multiplying $\mathscr{F}_3$ by $w$, the performance of the trained model is measured on the perturbed dataset.}
 \label{fig:measuring-performance}
\end{figure}

Fixing a class of interest ($c$) and a feature ($\mathscr{F}_i$) from the test set, we measure the stability of the model upon feature $\mathscr{F}_i$ based on two variables: a perturbation weight ($w$) inducing changes on all of the values of $\mathscr{F}_i$ and the performance score of the model after predicting labels for the test set with perturbation $w$ applied to $\mathscr{F}_i$. In other words, we multiply a weight $w$ to all values of feature $\mathscr{F}_i$ in the test set and use the model (trained with original training set) to measure the classifier performance after perturbation. Figure~\ref{fig:measuring-performance} illustrates this process for a dataset with $m$ features ($\mathscr{F}_1$, $\mathscr{F}_2$, ..., $\mathscr{F}_m$). Suppose a model was trained on these features, focused on the perturbation of $\mathscr{F}_3$ by an arbitrary weight $w$ (see $w \times \mathscr{F}_3$ highlighted). Without the perturbation on $\mathscr{F}_3$ ($w=1$), the model's performance is 0.98. Then, perturbing this feature with $w=1.3$, the model's performance decreases to 0.68.

%, as illustrated in Figure~\ref{fig:measuring-performance}. Notice that, with the perturbation on $\mathscr{F}_3$, the performance decreased from $0.98$ (with $w=1$) to $0.68$ (with $w=1.3$).

\begin{figure}[!htb]
 \centering % avoid the use of \begin{center}...\end{center} and use \centering instead (more compact)
 \includegraphics[width=\columnwidth]{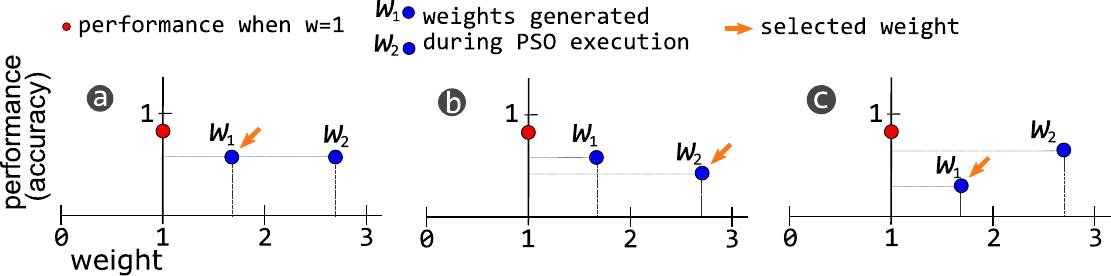}
 \caption{All of the scenarios for choosing weights according to their relationship to the models performance. The weight that induces the most decrease in performance is always selected---the closest one is selected for equal weights.}
 \label{fig:weights-explanation}
\end{figure}

Based on such an approach, changing the weight of important features (the ones that influence the classification) produces instability to the model by decreasing its performance. Thus, the higher is the decrease the more important is the feature associated with it. Notice that, since we multiply the perturbation weights by the features, these weights must have two characteristics: to induce a lot of decrease in the performance and to be close to one (i.e., when $w=1$ there is no perturbation). Figure~\ref{fig:weights-explanation} illustrates three possible scenarios for comparing perturbation weights and their respective model's performance: in the first case (\textbf{a}), two different weights induce the same accuracy, thus, we select $W_1$ which is the closest to one; in the second case (\textbf{b}), we select the second weight ($W_2$) since it induces the greater decrease; finally, we pick the select weight ($W_1$) in the third case (\textbf{c}) since it induces the greater decrease.

Notice that we try to minimize the performance and the perturbation at the same time. In other words, we try to induce more errors in the model using perturbation weights as close to one as possible. From these two objectives, minimizing performance is more important. For instance, to minimize the performance using the accuracy score as the performance measure, we can maximize Equation~\ref{eq:maximize-function}. Notice that $h$ is just a jargon of the PSO technique (the algorithm we used to maximize the function, explained in the following paragraphs), which is usually the name of the function that has to be optimized.

\begin{equation}
\label{eq:maximize-function}
h = |1 - Accuracy(y, \hat{y})|
\end{equation}

\noindent where $Accuracy(y, \hat{y})$ measures the accuracy of a model using the predicted labels $\hat{y}$ and the true labels $y$. Since $\hat{y}$ corresponds to the labels returned by prediction on a perturbed test set and recalling that perfect accuracy (when $y$ equals $\hat{y}$) assumes value $1$, maximizing Equation~\ref{eq:maximize-function} essentially means that we are looking for the weight $w$ that induces the most decrease in accuracy value. In other words, we try to minimize \textit{Accuracy($y$, $\hat{y}$)} by inducing errors through perturbations, which leads to maximizing Equation~\ref{eq:maximize-function}.

Equation~\ref{eq:maximize-function} depends only on $\hat{y}$, which depends on the $w$, thus, we have to define how labels $\hat{y}$ are predicted by the model after perturbation using $w$. First, the perturbation weights have to be separately defined for each pair of class and feature since the importance of features can be different from one class to another. For example, while a feature that describes the number of goals scored in a football game can be important for a forward player getting the prize of woman-of-the-match, it could have no importance for a goalkeeper getting the same prize. So that, to generate $\hat{y}$, the model ($f$) receives the matrix that joins disturbed and non-disturbed data samples, where the disturbing data samples corresponding to the instances of the same class as the analyzed ($c$). Equation~\ref{eq:generating-y} illustrates the idea, where $f$ denotes a classification model, $D^c$ denotes the sub-matrix whose data samples have class equals to $c$ and $D^{!c}$ denotes the sub-matrix whose data samples are not from class $c$---notice that $D$ corresponds to the test set. Finally,  $w$ $\times^i$ means the multiplication of all feature values of $\mathscr{F}_i$ by $w$.

\begin{equation}
\label{eq:generating-y}
\hat{y} = f([(w\times^i D^c )D^{!c}])
\end{equation}

Now, the question is how to find these weights. In this work, we use the PSO~\citep{Olsson2010} algorithm, a bio-inspired optimization algorithm that uses particles in the search space to optimize a given function. The PSO considers a set of particles used to cooperate in searching for the solution to an optimization problem. Each particle has its behavior and its group behavior (defined by a neighborhood). At the beginning of the algorithm, all particles receive a random position in the search space. Then, they are evaluated in each iteration according to an optimization function to verify if they are close or not to the solution of the problem. Notice that each particle update is influenced by the position of the best particle in its neighborhood and its past positions. During iterations, the best particles will lead the others throughout the search space.

Our implementation of the PSO algorithm (see Algorithm~\ref{alg:swarm-optimization}) tries to find the best weights that would maximize Equation~\ref{eq:generating-y} while considering the importance of these weights according to the scheme of Figure~\ref{fig:weights-explanation}. The first lines of the algorithm (lines 1-3) correspond to the initialization step of the particles ($X$ and $P$) and the velocity of movement ($V$) of these particles. Notice that $X$ and $P$ are initialized as matrices with ones since they represent the weights for each feature and we choose to initialize the search with no perturbation (i.e., with all weights equals one). These matrices have the dimensions of the number of particles ($p_m$) by the number of features ($m$). For a fixed number of iterations, we set as one all of the weights except for the weight corresponding to the feature in the analysis (lines $6$ and $7$)---notice that $X$ corresponds to the best particles' positions until the current iteration and $P$ corresponds to the positions of the particles in the current iteration. Setting to one all of the weights will induce change only in the feature of interest.  

\begin{algorithm}[]
\small
\caption{Finding minimum feature perturbation with PSO.}
\label{alg:swarm-optimization}
\begin{algorithmic}[1]
\Procedure{$\mathbf{PSO}$}{iterations, feature, $f$, $p_n$, $m$, $D$, $y$, $V_{min}$, $V_{max}$, $k$, $\chi$}
    
    \State X $\leftarrow$ $[1]_{p_n, m}$
    \State P $\leftarrow$ X
    \State V $\leftarrow$ $[V_{min} \leq random \leq V_{max}]_{p_n, m}$
	
	\While{ it $<$ iterations}
	    
	    \State $X[:, j \neq \text{feature}]$ $\leftarrow$ 1.0
	    \State $P[:, j \neq \text{feature}]$ $\leftarrow$ 1.0

	    \For{ each $p$ in $p_n$}
	        
	       \If{$h(X_p, D, y, f) > h(P_p, D, y, f)$ \textbf{or}\WRP $ h(X_p, D, y, f) \neq 0.0$ \textbf{and}\WRP $h(X_p, D, y, f) = h(P_p, D, y, f)$ \textbf{and}\WRP $\parallel 1 - X_{p, \text{feature}}\parallel < \parallel 1 - P_{p, \text{feature}}\parallel$} 
	           \State $P_p$ $\leftarrow$ $X_p$
	       \EndIf
	       
	       \State $g$ $\leftarrow$ $p$
	    
    	    \For{ each $j$ in neighborhood($p$)}
    	        \If{$h(P_j, D, y, f) > h(P_g, D, y, f)$ \textbf{or} \WRP $h(P_j, D, y, f) \neq 0.0$ \textbf{and} \WRP $h(P_j, D, y, f) = h(P_g, D, y, f)$ \textbf{and} \WRP $\parallel 1 - P_{j, \text{feature}}\parallel < \parallel 1 - P_{g, \text{feature}}\parallel$ }
    	           \State $g$ $\leftarrow$ $j$
    	       \EndIf
    	    \EndFor

	      %  \State $\phi_1$ $\leftarrow$ rand $\in [0, AC]$
	   %     \State $\phi_2$ $\leftarrow$ rand $\in [0, AC]$
    	    \State $V_p$ $\leftarrow$ $\chi\times[V_p + \phi_1\times(P_p - X_p) + \phi_2\times(P_g - X_p)]$
    	 %   \State $V_p \in [V_{min}, V_{max}]$ 
    	    
    	    \State $X_p$ $\leftarrow$ $X_p + V_p$
    	%    \State $X_p \in [X_{min}, X_{max}]$
	    
	    \EndFor
	  
	\EndWhile

\Return P
\EndProcedure
\end{algorithmic}
\end{algorithm}

The main part of the algorithm is used to find the best position for the particle $p$ ($P_p$) and the position of the best neighbor of $p$ ($P_g$). For each particle $p$ (line $8$), we use the function $h$ (see Equation~\ref{eq:maximize-function}) to verify if the current position of the particle $P_p$ is better than the previous position ($X_p$) or if $P_{p, feature}$ is closer than $X_{p,feature}$ to one when the induced performance by these two weights are equal---remember that from Figure~\ref{fig:weights-explanation} we choose the weight closest to one if the performance is the same. Having set the best particle ($p$) in line $10$, we initialize the best particle in the group ($g$) as being $p$ as well. Lines 12-14 do essentially the same thing as discussed for lines 8-10, however, looking only to the neighborhood of $p$. In the end, $p$ contains the index of the best particle and $g$ contains the index of the best particle in the group. 

The final part of the iteration of the algorithm consists of updating $X_p$ (the matrix row storing the best position of the particle $p$) based on the positions of the best particle ($p$) and the best in the group ($g$). Line 15 uses such information to compute the new velocity vector based on the previous velocity ($V_p$) and the position of the particle ($X_p$). According to the PSO algorithm, $\phi_1$ and $\phi_2$ are random vectors and the multiplication $\phi_1\times(P_p-X_p)$ corresponds to the individual behavior of the particle while $\phi_2\times(P_g-X_p)$ corresponds to the social behavior of the particle. In other words, these two equations mean to move particle $p$ based on its current best value and based on the state of its neighborhood. Finally, the parameter $\chi$ is the constriction coefficient and it helps the algorithm in the convergence, we set $\chi = 0.729$ since it is a well-established value in the literature.

Although the algorithm has a lot of parameters, the literature shows that $V_{min}=-1$ and $V_{max}=1$ works fine, so we fixed them. The neighborhood was defined as $k=1$ following a ring pattern, i.e., for each particle $p$, its set of neighbors correspond to the indices $i$, $p-k\leq i\leq p+k$, $i\neq p$. Finally, at each iteration, we normalize set $X$ values to be inside the range $[0, 10]$.

After execution, |$p_m$| weights in the $X_{k, feature}$ column vector are available for the feature in analysis to compute the feature's importance. Thus,  the mean of the absolute difference between the weights and one $\sigma_w = \frac{1}{p_m}\sum\limits_{j=1}^{p_m}|w_j - 1|$ and the mean of the scores when predicting using each weight minus the score with no perturbation $\sigma_s = \frac{1}{p_m}\sum\limits_{j=1}^{p_m}|S(w_j) - S(1)|$ is taken. Then, the importance for each feature is computed as follows:

\begin{equation}
\label{eq:swarm-importance}
I_i = \sigma_{s_i} \times (X_{max} - [(\frac{\sigma_{w_i}}{max(W_{\sigma_{s_i}})})/X_{max}])
\end{equation}

\noindent where $W_{\sigma_{s_i}}$ is the set of weights with respective score same as $\sigma_{s_i}$. Essentially, with Equation~\ref{eq:swarm-importance} we derive a number that depends on the mean weight $\sigma_{w_i}$ and its contribution to the model's performance. In this case, with $\sigma_{s_i}$ multiplied by $X_{max}$ we can order the importance based on accuracy, then, $\sigma_{s_i}$ multiplied by the normalized weight (among those with same mean score $\sigma_{s_i}$) results in an ordering for the features with the same $\sigma_{s_i}$. After computing the importance of each feature, they can be ordered in a decreasing way. Table~\ref{tab:importances} exemplifies such an ordering for the \textit{Iris} dataset, notice that the features are arranged as if they were ordered based on $\sigma_w$ and $\sigma_s$, consecutively. Notice that, the higher is $I$ the more important is the feature.

\begin{table}[!htb]
\centering
\caption{Feature importance ($I$) defined by the mean weight ($\sigma_w$), and the performance weight ($\sigma_s$).}
\label{tab:importances}
\begin{tabular}{lccc}
\toprule
Feature           & $\sigma_w$ & $\sigma_s$ & $I$       \\ 
\midrule
petal width (cm)  & $0.37515$  & $0.43333$  & $4.29485$ \\ 
petal length (cm) & $0.42242$  & $0.43333$  & $4.29000$ \\ 
sepal length (cm) & $0.03375$  & $0.00000$  & $0.00000$ \\ 
sepal width (cm)  & $0.31219$  & $0.00000$  & $0.00000$ \\
\bottomrule
\end{tabular}

\end{table}

\section{Visualization Design}
\label{sec:visualization-design}

Using the methodology described in the previous section, we defined several design requirements (DR) to help with the interpretation of classification models using the PSO algorithm. The visualization is based on the visual inspection of the PSO algorithm execution with carefully chosen visual variables to help interpret classification results. Using the information generated by the PSO algorithm and other requirements to interpret a model's decision, we want to be able to visualize:

\begin{itemize}
    \item \textbf{DR1}: the weights that most influence the classifier decision;
    \item \textbf{DR2}: the strength of influence of a feature;
    \item \textbf{DR3}: the confusion of the classifier based on different perturbation weights;
    \item \textbf{DR4}: the distribution of values for the features;
    \item \textbf{DR5}: the PSO execution;
    \item \textbf{DR6}: the similarity among the classes in the dataset.
\end{itemize}

To accomplish these design requirements (DR), we follow a strategy to create the visualization tool centered on the PSO execution. The interpretation of the classifier's decisions consists of inspecting the feature perturbations generated by PSO particles. In the following sections, we explain the detailed and the summary visualization for a class of interest. Then, we show how to measure the similarity between classes in terms of classifier confusion.

\subsection{Detailed View}

In the PSO algorithm, the particles assume different values (weights) during iterations. These weights correspond to the perturbations we want to find to interpret a model's prediction. When we multiply a weight $w_i^j$ to a feature $\mathscr{F}_k$, we can assess how much the perturbation $w_i^j$ induces change on the classification performance---notice that $i$ represents the particle index and $j$ represents the current iteration of the PSO algorithm. With such an idea in mind, we can use graphical variables to encode the relationship between the perturbation ($w_i^j$) and the feature. Notice that the perturbation consists of multiplying $w_i^j$ by $\mathscr{F}_k$ $(w_i^j\times \mathscr{F}_k)$, in other words, multiplying $w_i^j$ by the column $k$ of the test set.

\begin{figure}[!htb]
 \centering % avoid the use of \begin{center}...\end{center} and use \centering instead (more compact)
 \includegraphics[width=\columnwidth]{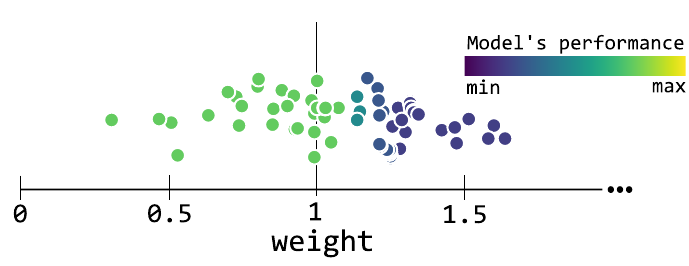}
 \caption{Encoding PSO execution. Each circle corresponds to a value (weight) assumed by a particle during PSO's execution, while the colors encode the model performance when a feature is multiplied by those weights. Here, color-coding indicates that the classifier is sensitive when the feature values increase.}
 \label{fig:encoding-strategy}
\end{figure}

To visualize the perturbations and the result on the performance measure after a model's prediction, we encode the weights as circles in a horizontal axis. At the same time, a color scale represents the change in the model's performance, as illustrated in Figure~\ref{fig:encoding-strategy}. Such an encoding shows consistency with PSO execution since we move the weights from the initial position (close to one) to positions with much perturbation to the model. Each circle corresponds to a weight assumed by a particle during PSO execution---notice that the $y$-axis does not have any meaning.

\begin{figure*}[!htb]
 \centering % avoid the use of \begin{center}...\end{center} and use \centering instead (more compact)
 \includegraphics[width=\textwidth]{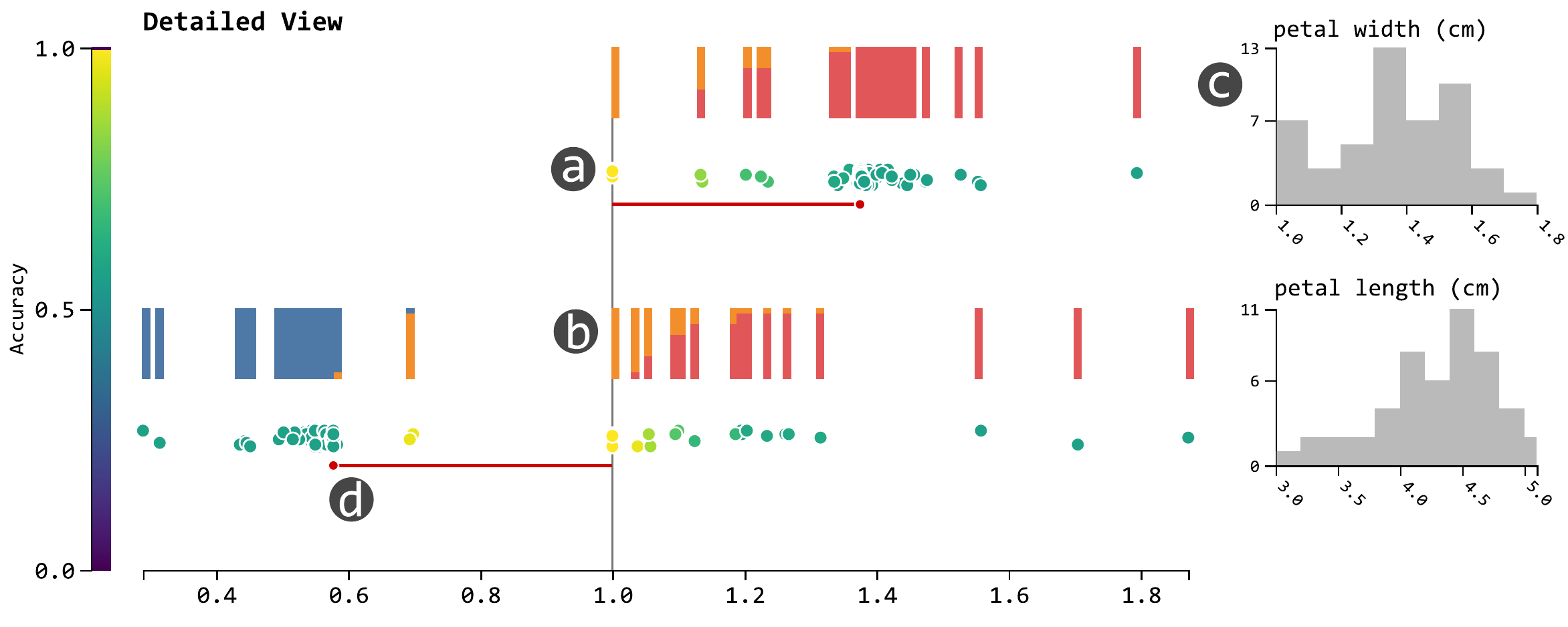}
 \caption{Detailed View for classifier interpretation using PSO. Using the PSO execution encoding (\textbf{a}) as explained in Figure~\ref{fig:encoding-strategy} we visualized how the weights affect on model's prediction. The stacked bar charts encode the classification proportion inside a weight range ($[a, a+\epsilon]$) to assist in the classifier's confusion analysis. Distribution plots (\textbf{c}) of the feature values for the class of interest also help with the interpretation of the results. Finally, the horizontal red segment (\textbf{d}) also assists in the identification of the best weight for the feature concerning the definition of our optimization function.}
 \label{fig:detailed-view}
\end{figure*}

Using the visual encoding of Figure~\ref{fig:encoding-strategy} we can visualize the weights that most influence the classifier decision (DR1) by comparing the position (distance to one) and the induced model's performance. The circles and their color also help to understand the strength of influence (DR2) and the PSO execution (DR5) throughout iterations based on the weights assumed by the particles. However, other additional information could help to interpret the classifier decisions and learning processes. Firstly, the distribution of values (DR4) for the feature $\mathscr{F}_k$ helps to contrast the classifier performance with the pattern seen in data samples for a particular class. Lastly, for the non-binary classification problem, it is interesting to know how different weights influence the classifier's confusion with other classes (DR3). Figure~\ref{fig:detailed-view} shows information for two features. Apart from distribution plots showing the feature values (\textbf{c}), we encode the confusion of the classifier after perturbation using proportions of the confused classes (\textbf{b}) (indicated by color hue)---notice that higher bars show a greater number of data samples classified as that specific class. The circles in (\textbf{a}) encode the information already discussed and exemplified with the scheme of Figure~\ref{fig:encoding-strategy}. Finally, the horizontal red segment (\textbf{d}) shows the best weight (among all of the weights generated by the particle) for the feature concerning the definition of our optimization problem, that is, the minimum weight that induces the higher loss to the model's performance.

One important thing to mention is that the proportion displaying the confusion is not computed using one weight but a range of weights inside a window. More specifically, we divide the weight space in windows of length $\epsilon$ and use all weights inside each window to calculate the proportion. Since each perturbation weight generates $m$ predictions---where $m$ is the number of data samples in the test set---, we compute the mean of classification for each class to derive the proportion of the bars in the stacked bar chart.

\subsection{Summary View}

The visualization design presented in the previous section suffers from visual scalability issues when analyzing a large dataset (in terms of dimensionality). So, we designed a summary version of the information presented in the detailed view where users can choose which features to explore throughout the interaction. More specifically, in the \textit{Summary View}, we show the relationship between perturbation weights and the model's performance.

\begin{figure}[!htb]
 \centering % avoid the use of \begin{center}...\end{center} and use \centering instead (more compact)
 \includegraphics[width=\columnwidth]{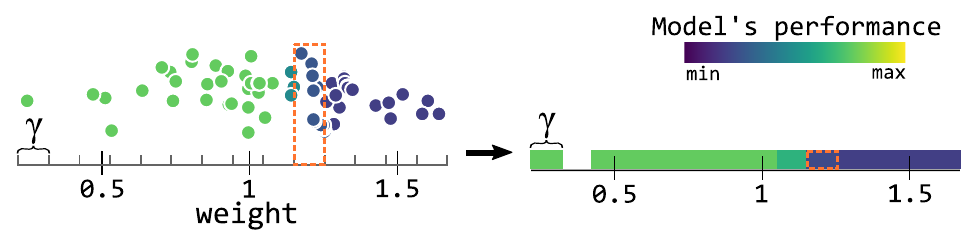}
 \caption{Aggregation of particle weights. Using a range ($[a, a+\gamma]$), the mean of the weights inside the range is encoded by the color scale. The area highlighted in orange shows how different weights are aggregated in a single block.}
 \label{fig:summarization-scheme}
\end{figure}

To summarize the information contained in the \textit{Detailed View}, we use a variable $\gamma$ to create windows on the weight axis, as illustrated in Figure~\ref{fig:summarization-scheme}. Then, each feature weight inside a window induces a performance score ($ps$) to the model being analyzed. Since the mean of performance score inside a window still lies inside the score range (from $0$ to $1$ for accuracy), windows are color-coded based on the same color-scale used for the \textit{Detailed View}. Notice that for some weights, there is no influence on the model's prediction (see the performance for weights lower than one in Figure~\ref{fig:summarization-scheme}) while there is an upper bound for how much change on the performance a feature perturbation can induce (see the performance for weights greater than approximately $1.25$).

\begin{figure}[!htb]
 \centering % avoid the use of \begin{center}...\end{center} and use \centering instead (more compact)
 \includegraphics[width=\columnwidth]{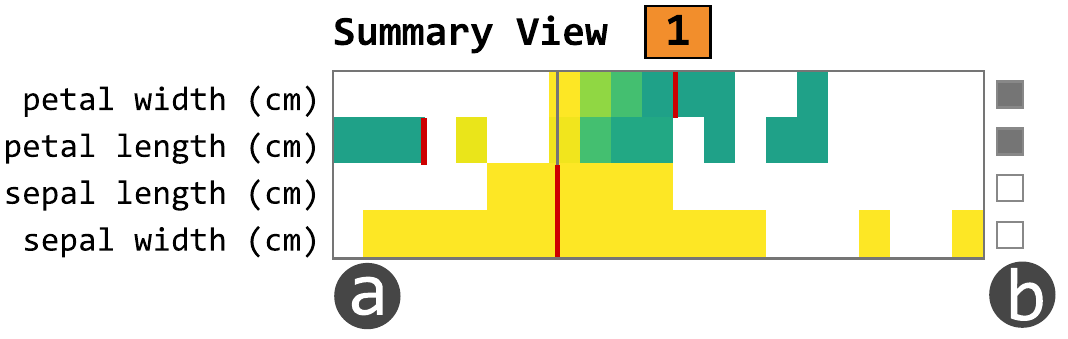}
 \caption{Summary View showing all information about PSO execution in an aggregated way for class $1$ (orange). The color of the blocks (\textbf{a}) encodes classifier performance (e.g. accuracy) and the red segments show the best weight for each feature. The grey vertical segment shows the weight with no perturbation ($w = 1$). Finally, the boxes on the right of each feature show which features are being inspected at the moment (grey - selected, while - unselected).}
 \label{fig:summary-view}
\end{figure}

Figure~\ref{fig:summary-view} shows the results of summarizing information using the discussed strategy. Besides the performance of the classifiers inside the windows defined using $\gamma$, the best weight (as the one for the \textit{Detailed View}) is shown through a vertical segment in red. Lastly, the boxes on the right of each axis show which features are being currently inspected on the \textit{Detailed View}. Users can toggle the boxes to show (when the boxes are filled with gray) or hide (when the boxes are filled with white) feature information.

\subsection{Similarity View}

Besides assessing how the perturbations affect a model's performance based on a local perspective, a global view, where we understand how the classification of class $a$ is confused with class $b$, can be helpful to guide users on the detailed inspection. Notice that confusion is not reflexive, that is, the confusion from $a$ to $b$ is not the same as the confusion from $b$ to $a$.

Let us define the similarity between two classes $a$ and $b$ (from $a$ to $b$) as a measure of confusion of the classifier between these classes. Firstly, we must give higher weight to the data points confused with $b$ when the weight induced to such confusion is closer to one. This means we give more interaction to the classes in which little perturbation induces more confusion. So that, the similarity from class $a$ to class $b$ for feature $i$ is defined as it follows:

\begin{equation}
\label{eq:interaction-i}
I_i^{a\rightarrow b} = \frac{ \sum\limits_{w \in W}(W_{max} - w)\times |\lbrace k \in f(w\times D_{*, i}): k = b\rbrace| }{ \sum\limits_{w \in W}(W_{max} - w)\times |D| }
\end{equation}

Equation~\ref{eq:interaction-i} ranges from $0$ to $1$, where $1$ means that all of the test data samples were classified as $b$. Notice that $|\lbrace k \in f(w\times D_{*, i}): k = b\rbrace|$ denotes the number of data points classified as $b$ after multiplying $w$ to the column corresponding to feature $i$. $W_{max}$ corresponds to the greatest value that a particle can assume in the PSO algorithm (here, we set $W_{max} = 10$). Using that equation, we can define the similarity from $a$ to $b$ as the mean of similarities of each feature, as shown in Equation~\ref{eq:interaction}.

\begin{equation}
\label{eq:interaction}
I^{a\rightarrow b} = \dfrac{1}{m}\sum\limits_{i=1}^{m}I_i^{a\rightarrow b}.
\end{equation}

\begin{figure}[!htb]
 \centering % avoid the use of \begin{center}...\end{center} and use \centering instead (more compact)
 \includegraphics[width=\columnwidth]{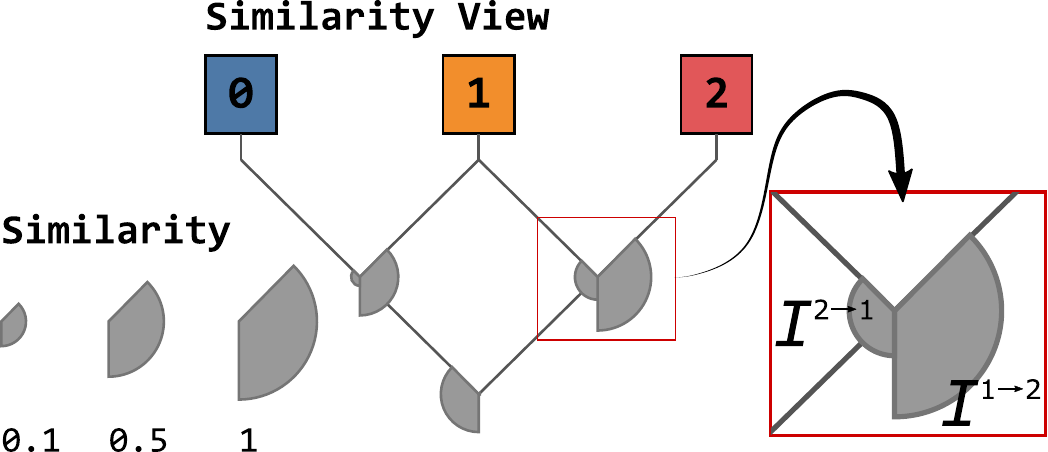}
 \caption{Similarity View encodes the relationship between each pair of classes. The similarity from class $a$ to $b$ ($I^{a\rightarrow b}$) is visualized by the outer radius of a radial layout segment. Notice that such an encoding follows the direction of the interaction, as indicated by the red square.}
 \label{fig:interaction-view}
\end{figure}

To visualize the similarity information among all classes, we use a node-link layout together with an encoding inspired on the UpSet~\citep{Lex2014} visualization. As illustrated in Figure~\ref{fig:interaction-view}, every pair of combination is encoded by a cross on the lines of the classes. Then, to communicate the interaction $a$ to $b$ for every pair of ($a, b$), we use the outer radius of an arc positioned on the crossing between $a$ and $b$. Notice in Figure~\ref{fig:interaction-view}, the encoding for interaction between two classes follows the direction of the source class to the target, as demonstrated for $I^{1\rightarrow 2}$ and $I^{2\rightarrow 1}$. For instance, the figure shows the example of a Similarity View for the \textit{Iris} dataset. Notice that, as it is very understanding about this well-known dataset, two classes interact the most and are responsible for errors in the classification.

\section{Use cases}
\label{sec:use-cases}

In this section, we use the feature perturbations generated by PSO to interpret the model's behavior upon the classification of several datasets. All of the experiments were performed with a computer with the following configuration: Intel (R) Core(TM) i7-8700 CPU @ 3.20 GHz, 32GB RAM, Windows 10 64 bits. Since the focus is on the interpretation of the results rather than the performance of the classifier, all of the hyper-parameters and details are described in the \textbf{Supplementary File}.

\subsection{Vertebral Column}

\begin{figure}[!htb]
 \centering % avoid the use of \begin{center}...\end{center} and use \centering instead (more compact)
 \includegraphics[width=0.7\columnwidth]{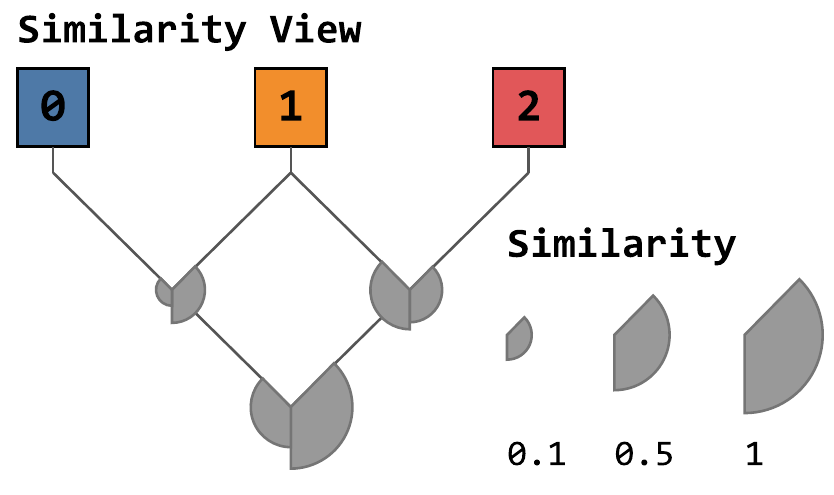}
 \caption{Similarity View shows there is a lot of interaction between classes $0$ (blue) and $2$ (red). However, class $1$ (orange) seems to have very distinctive characteristics.}
 \label{fig:vertebral-similarity}
\end{figure}

In this first use case, we interpret the XGBoost Classifier~\citep{Chen2016} applied to the \textit{Vertebral Column}~\citep{Dua2019} dataset, composed by $310$ instances described by six biomechanical features derived from the shape and orientation of the pelvis and lumbar spine: \texttt{pelvic incidence}, \texttt{pelvic tilt}, \texttt{lumbar lordosis angle}, \texttt{sacral slope}, \texttt{pelvic radius}, and \texttt{grade of spondylolisthesis}. The dataset is divided into three classes: class \img{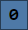} for patients with Hernia, class \img{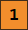} for patients with Spondylolisthesis -- a disturbance of the spine in which a bone (vertebra) slides forward over the bone below it, and class \img{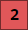} for normal patients. Figure~\ref{fig:vertebral-similarity} shows the interaction among these three classes. From the Similarity View, we can understand that there is a lot of confusion between classes \img{class0} and \img{class2} induced by the perturbation weights generated by the PSO execution, which indicates that it is a difficult task to determine whether a data sample belongs to  either of these two classes. On the other hand, class \img{class1} seems to have very distinctive features from the other two classes, where the confusion is mostly present in the backward form ($I^{0\rightarrow 1}$ and $I^{2\rightarrow 1}$).

\begin{figure}[!htb]
 \centering % avoid the use of \begin{center}...\end{center} and use \centering instead (more compact)
 \includegraphics[width=\columnwidth]{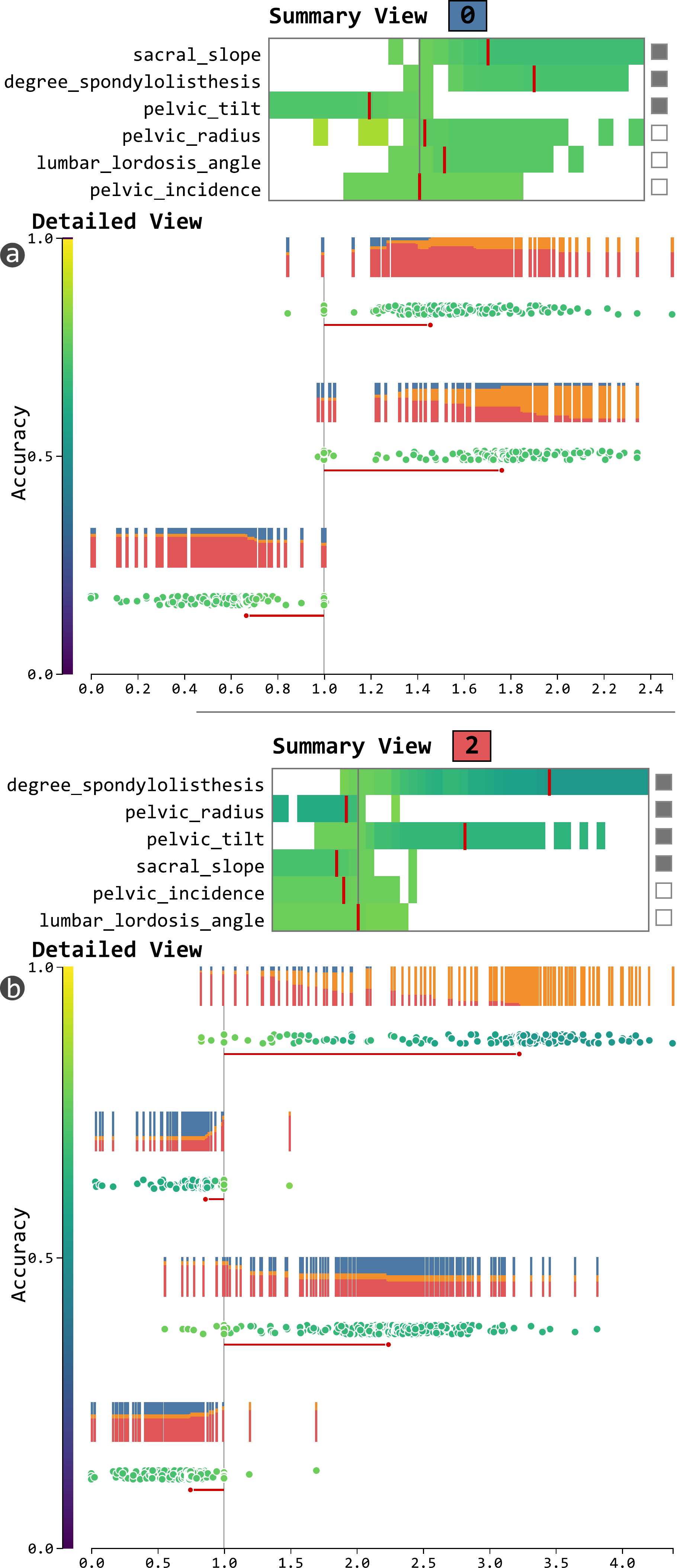}
 \caption{Detailed View for both patients with Hernia (\textbf{a}) and normal patients (\textbf{b}) shows that except for \texttt{degree of spondylolisthesis}, these two classes are highly confused between each other by the classifier.}
 \label{fig:vertebral-class0-class2}
\end{figure}

To inspect how the classes \img{class0} and \img{class2} present such complexity to the classifier, Figure~\ref{fig:vertebral-class0-class2} shows the summary result of all features in the dataset and detail of only the most distinctive features for these two classes. The first thing to notice is that the classifier confuses both of the classes with class \img{class1} when \texttt{degree\_spondylolisthesis} increases -- as we show later, the degree of spondylolisthesis is the most determinant feature for patients with Spondylolisthesis. Further that, we can identify that those normal patients (class \img{class2}) present lower values for pelvic tilt and higher values for sacral slope and pelvic radius, that is, see how the confusion from class~\img{class0} to class~\img{class2} increases when the feature \texttt{pelvic tilt} is multiplied by a number below one or when the feature \texttt{sacral slopse} is multiplied by a number higher than one in Figure~\ref{fig:vertebral-class0-class2}\textbf{a}. On the other hand, the values for these features are found to be the opposite in patients with Hernia~\citep{Labelle2005, Roussouly2011, Huec2011}. We can see that by analyzing the changes induced by the perturbations generated with the PSO algorithm, we could understand a lot of characteristics of the dataset together with the classifier decisions. Here, while the classifier could learn the parameters to be more confident when the features are pushed to the limit (higher or lower values), it seems that the classifier needs improvement to correctly classify data samples that share class boundaries -- when feature values are similar.

\begin{figure}[tb]
 \centering % avoid the use of \begin{center}...\end{center} and use \centering instead (more compact)
 \includegraphics[width=\columnwidth]{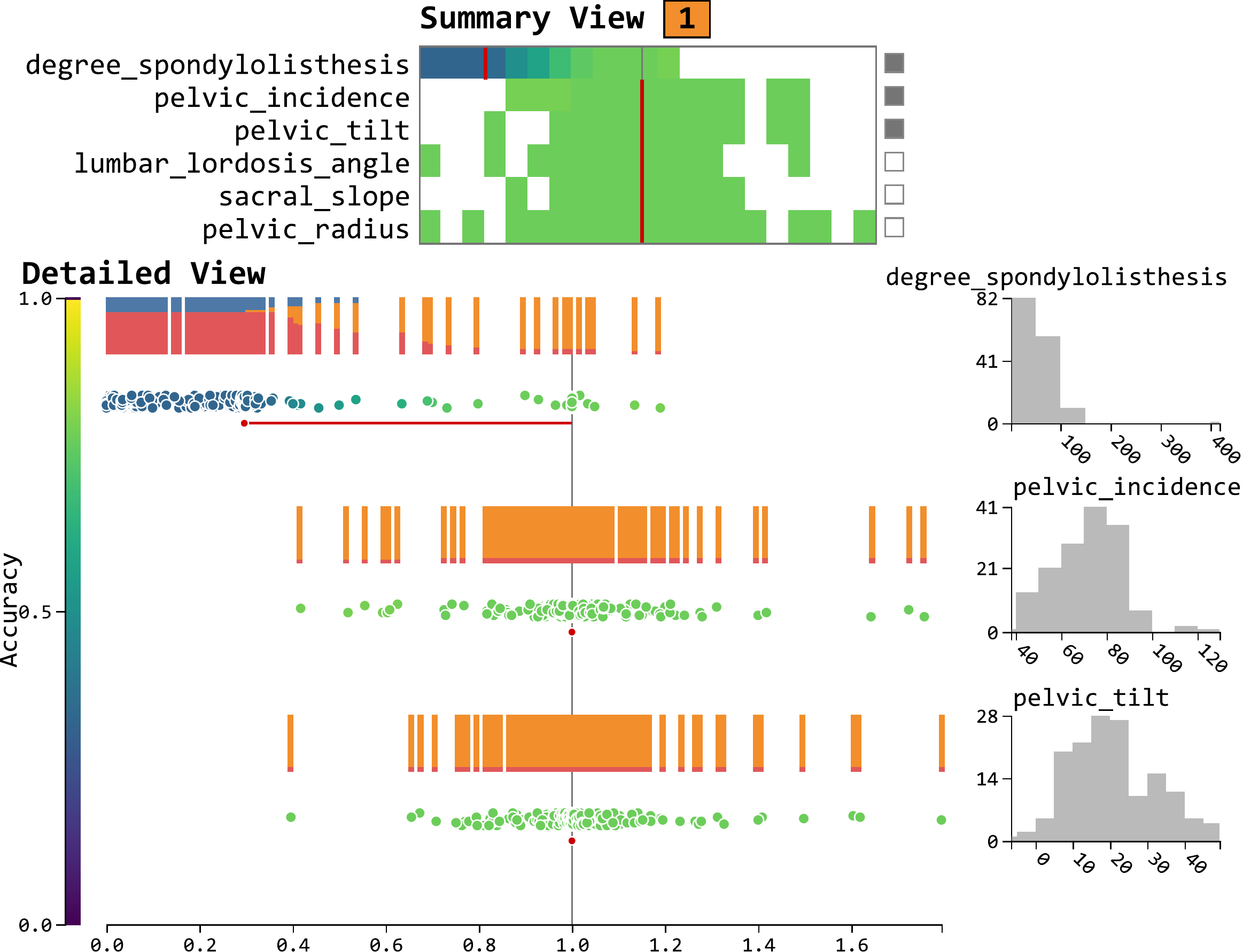}
 \caption{For patients with Spondylolisthesis, the classifier assigned the feature \texttt{degree of spondylolisthesis} as the most important feature, meaning that it is responsible for correct classification of such a class.}
 \label{fig:vertebral-class1}
\end{figure}

The class \img{class1}, as shown in Figure~\ref{fig:vertebral-class1}, presents different results only when the perturbations are applied to the degree\_spondylolisthesis feature. Notice that how the results are consistent with the Similarity Plot, in which there are very low similarity going from class \img{class1} to the others.

Now, we must verify if the results of the classifier explained by the PSO perturbation weights make sense. That is, to verify if the patients with Spondylolisthesis, the degree of spondylolisthesis are known to be greater. Interestingly, the degree of spondylolisthesis is the most important factor to determine if a patient has Spondylolisthesis or not~\citep{Labelle2005}. This makes sense if, with a higher degree, a vertebra bone presents more deviation from a bone below it -- which constitutes the Spondylolisthesis. Here, we see that the classifier used \texttt{degree\_spondylolisthesis} to induce a separation between class~\img{class1} and the other two classes~\img{class0} and~\img{class2}.

\subsection{Heart disease}

In this second case study, we use our approach to inspect and understand the features used by Random Forest Classifier~\citep{Breiman2001} to differentiate between patients with healthy and unhealthy hearts. The dataset contains two classes: healthy hearts (\img{class0}) and unhealthy hearts (\img{class1}). Each data sample is described by the following features: \texttt{age}, \texttt{sex} (1 - male; 0 - female), \texttt{chest pain type}, \texttt{resting blood pressure}, \texttt{serum cholesterol} (in mg/ml), \texttt{fasting blood sugar}, (> $120$ mg/ml), \texttt{resting electrocardiographic results} (values $0, 1, 2$), \texttt{maximum heart rate achieved}, \texttt{exercise induced angina}, \texttt{oldpeak} = ST depression induced by exercise relative to rest, the \texttt{slope of the peak exercise} ST segment, \texttt{number of major vessels} (0-3), and \texttt{thal} (3 - normal, 6 - fixed defect, 7 - reversable defect). Since there are only two classes, the Summary View does not add much information to the analysis, thus, we omitted it for this case study.

\begin{figure}[!htb]
 \centering % avoid the use of \begin{center}...\end{center} and use \centering instead (more compact)
 \includegraphics[width=\columnwidth]{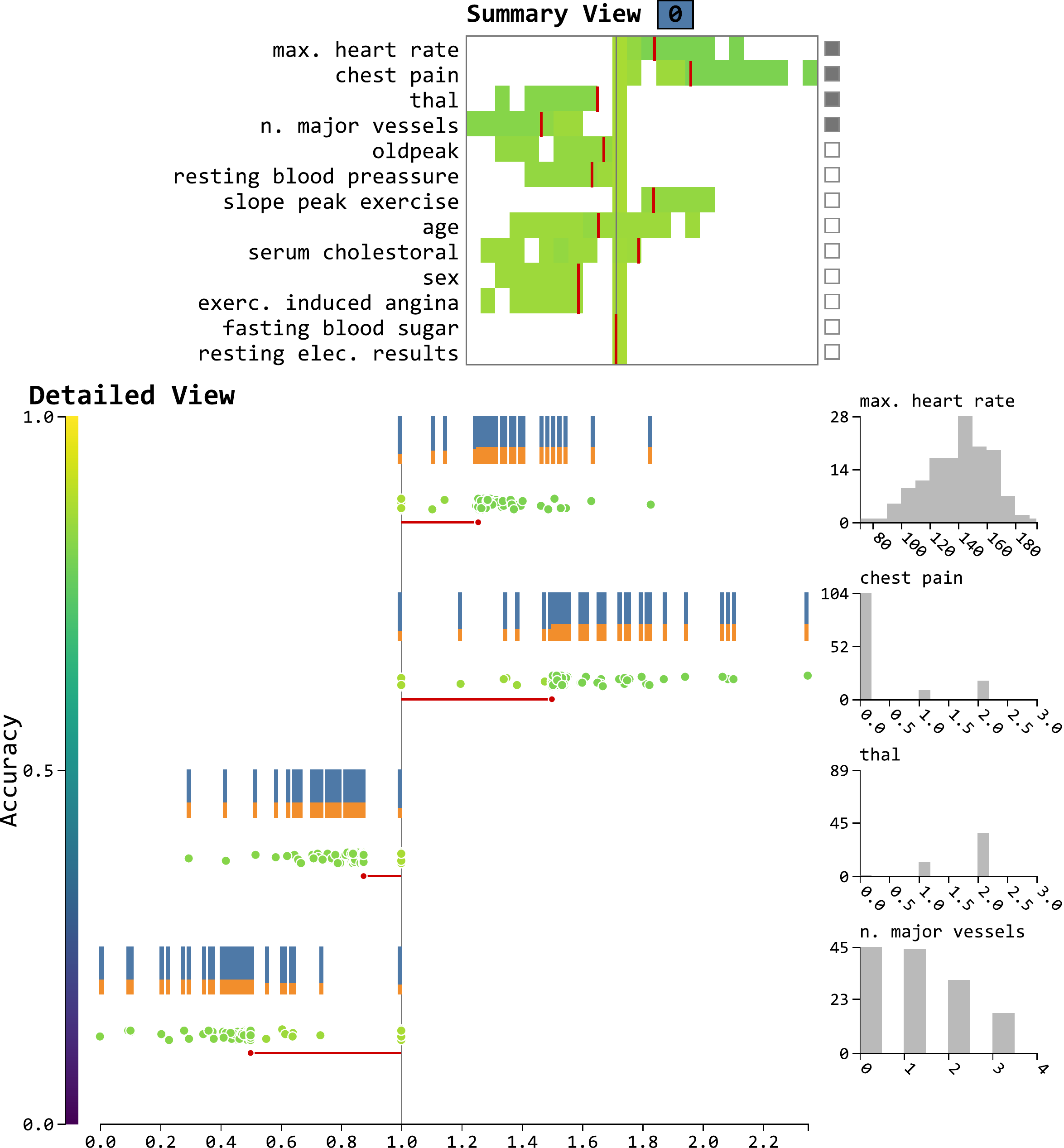}
 \caption{Detailed View for patients with healthy hearts. Our technique correctly shows that the classifier was able to learn that a healthy heart has a moderate heart rate while lower severity in chest pain.}
 \label{fig:heart-class0}
\end{figure}

Figure~\ref{fig:heart-class0} shows the summary and detailed (for some features) of the importance given by the classifier to classify data samples as healthy hearts. The most important feature, that according to the classification model constitutes a healthy heart, is \texttt{max. heart rate}. In this case, healthy hearts are seen by the classification model as the ones with moderate rate beat -- when the heartbeat gets higher, the model starts to confuse with unhealthy hearts. Then, the \texttt{chest pain}, defined by increasing values related to the severity (from $0$ to $3$), constitutes the second most important feature. As learned by the algorithm, healthy hearts present lower levels of severe pain (see the distribution plot of \texttt{chest pain}). The third most influential feature, \texttt{thal}, corresponds to an inherited blood disorder (Thalassemia) that causes one's body to have less hemoglobin than normal. Interestingly, heart problems (such as congestive heart failure and abnormal heart rhythms) can be associated with Thalassemia~\citep{Jameson2019, Thalassemias2019}. Finally, the model also gave importance to the \texttt{number of major vessels} feature. This feature indicates how many arteries are visible after a special dye (fluoroscopy) is injected into the blood vessels of the heart. Unlike the previously discussed feature (\texttt{thal}), there is a consistency to the algorithm as defining healthy hearts the one with a higher number of visible vessels.

\begin{figure}[!htb]
 \centering % avoid the use of \begin{center}...\end{center} and use \centering instead (more compact)
 \includegraphics[width=\columnwidth]{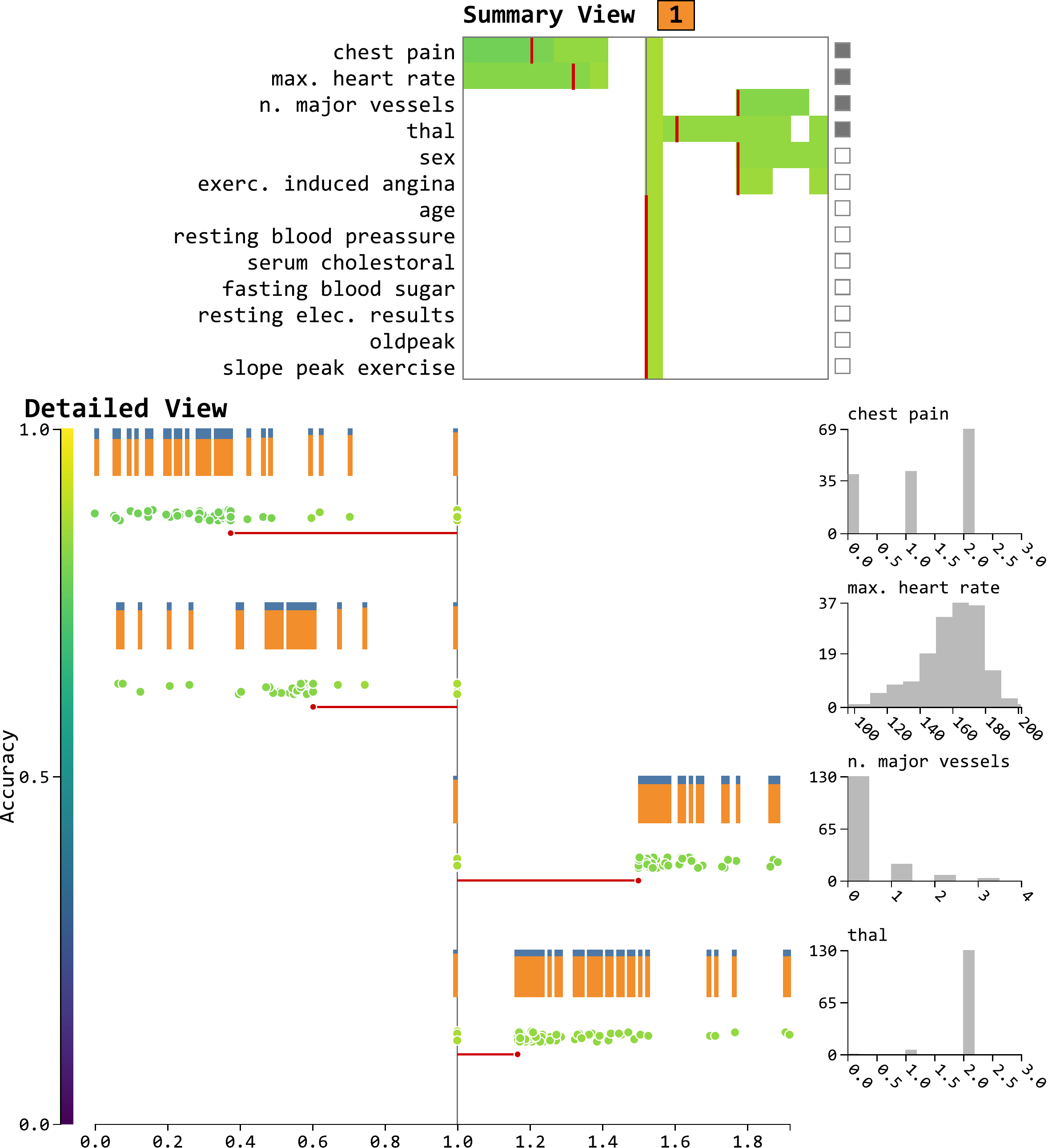}
 \caption{As discussed for patients with healthy hearts, our approach consistently shows that the classifiers assigned the data samples as unhealthy when they have higher \texttt{max. heart rate} and severe \texttt{chest pain}. Besides that, unhealthy hearts also seem to present a lower number of visible vessels.}
 \label{fig:heart-class1}
\end{figure}

The pattern perceived for unhealthy hearts (class~\img{class1}) is nearly the opposite of the ones perceived for healthy hearts, as shown in Figure~\ref{fig:heart-class1}. Although the four most important features of class~\img{class1} are the same four most important features of class~\img{class0}, there is a change in the ordering. For the case of \texttt{chest pain}, the feature weights are lower than one since unhealthy hearts present more severe chest pain, encode by greater values. The second most important feature, \texttt{max. heart rate} shows how unhealthy hearts present a beat rate greater than healthy hearts. Then, the narrowing of blood vessels, which is related to the number of major vessels, is usually due to arteriosclerosis, a common arterial disease in which increased areas of degeneration and cholesterol deposit plaques form on the inner surfaces of the particles blocking the blood flow~\citep{Gersh2000, Khatibi2010}. Finally, one must understand the organization of the dataset and the field in which the problem is inserted before taking any particular assumption. That is, besides preventing analysts from defining erroneous hypothesis, carefully analyze the results returned by our algorithm will help one to understand the decisions of the model and the structures of the dataset, such as data mismatch. For example, after receiving a high score in a classification task, one could analyze and understand if there is bias in the features learned by the models.

\subsection{Diabetes}

In this final case study, we interpret CatBoost~\citep{Dorogush2018} classifier's prediction on a dataset containing $768$ data samples of patients described by eight medical features concerning the presence of diabetes: $268$ data samples of patients with diabetes (class~\img{class0}) and $500$ data samples of non-diabetic patients (class~\img{class1}). The features used for classification are: \texttt{preg} (number of times pregnant), \texttt{plas} (plasma glucose concentration after 2 hours in an oral glucose tolerance test), \texttt{pres} (diastolic blood pressure), \texttt{skin} (triceps skin fold thickness (mm)), insulin (2-hour serum insulin (mU/ml)), \texttt{mass} (body mass index (weight in kg/(height in m)$^2$), \texttt{pedi} (diabetes pedigree function), \texttt{age} (age in years). As we discussed for the previous case study, we do not rely on the similarity view to extract information of this dataset since it consists in a binary classification.

\begin{figure*}[!htb]
 \centering % avoid the use of \begin{center}...\end{center} and use \centering instead (more compact)
 \includegraphics[width=0.8\textwidth]{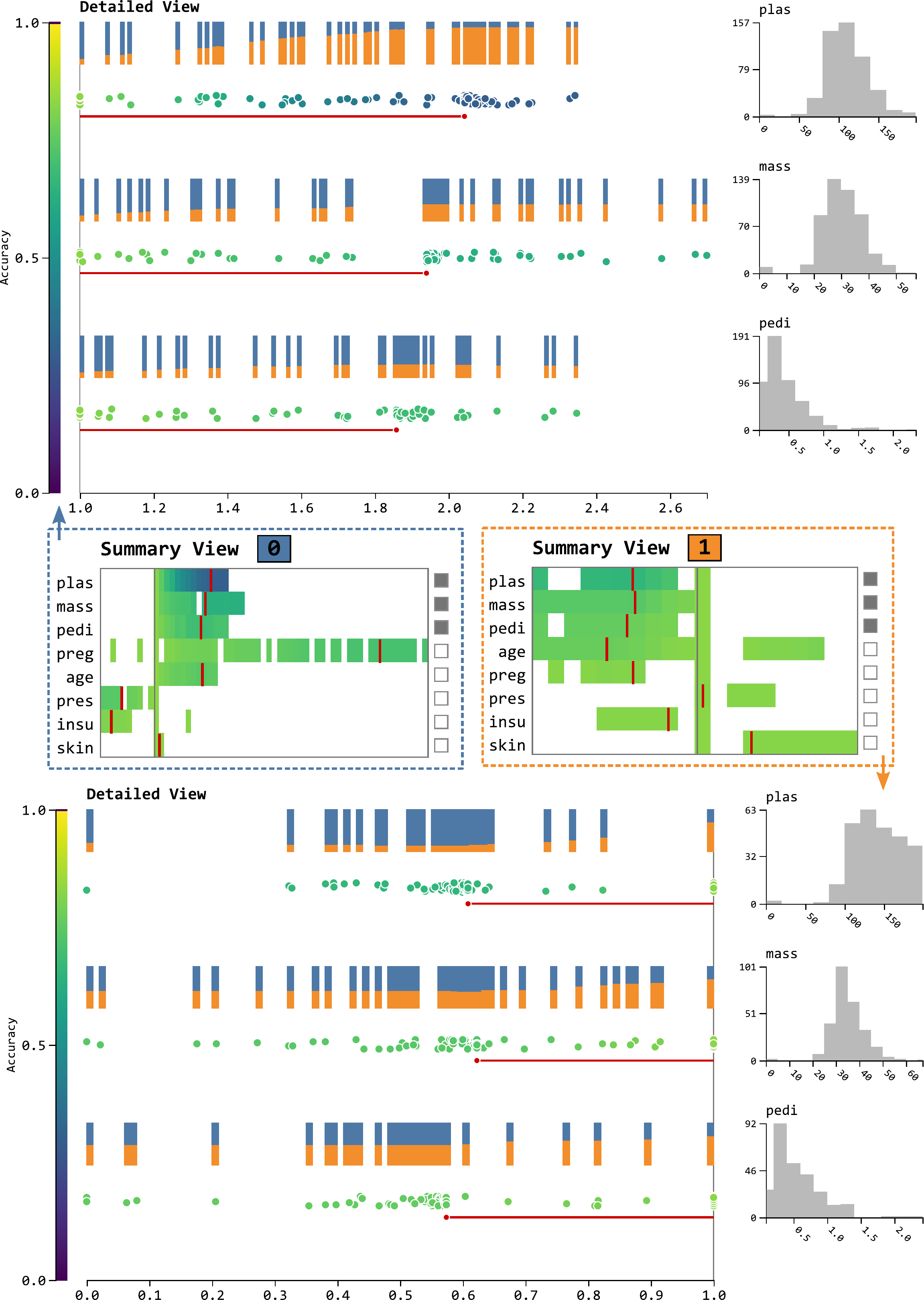}
 \caption{Detailed View showing that the classifier gave the same importance ordering for both non-diabetic and diabetic patients. However, we can see that the weights generated by the PSO algorithm are consistent with classes. For instance, non-diabetic patients (class $0$ - blue) need to have \texttt{plas} and \texttt{mass} increased to be classified as diabetic.}
 \label{fig:diabetes-class0-class1}
\end{figure*}

Figure~\ref{fig:diabetes-class0-class1} shows the result for both classes: patients with diabetes~\img{class0} and patients without diabetes~\img{class1}. Here, we focus on the three most important features returned by our technique for both of the classes, in which three importance ordering is the same: plasma glucose concentration (\texttt{plas}), body mass index (\texttt{mass}), diabetes pedigree function (\texttt{pedi}). Recalling that plasma glucose concentration (or simply blood sugar level) is a well-known indicator of prediabetes when the levels are high~\citep{AbdulGhani2010}, our approach consistently shows how a healthy patient (class~\img{class0}) must increase his blood sugar levels to become diabetic---notice that it consistently identified this feature as the most important one. Looking at the distribution of values for \texttt{plas} for patients without diabetes, there is a concentration of around $100$. Further, the second most important feature also shows that increased body fat will result in patients with a higher probability of having diabetes. Interestingly, such a result is consistent with the literature since an increase in body fat is generally associated with an increase in the risk of metabolic diseases such as Type $2$ diabetes mellitus~\citep{Bays2007}. Finally, the diabetes pedigree function feature (\texttt{pedi}) also illustrates the applicability of our visualization design and method of model explanation. Since such a feature measures the likelihood of diabetes based on family history, it is clear that higher pedigree function valuers will induce more chances for patients to be classified as diabetic.

Interestingly, looking at the perturbations of class~\img{class1}, the algorithm consistently imposed weight that would result in the levels of features shown by healthy patients. That is, normal levels for \texttt{plas}, \texttt{mass}, and \texttt{pedi}.

\section{Numerical evaluation}
\label{sec:numerical_evaluation}

\begin{figure}[tb]
 \centering % avoid the use of \begin{center}...\end{center} and use \centering instead (more compact)
 \includegraphics[width=\linewidth]{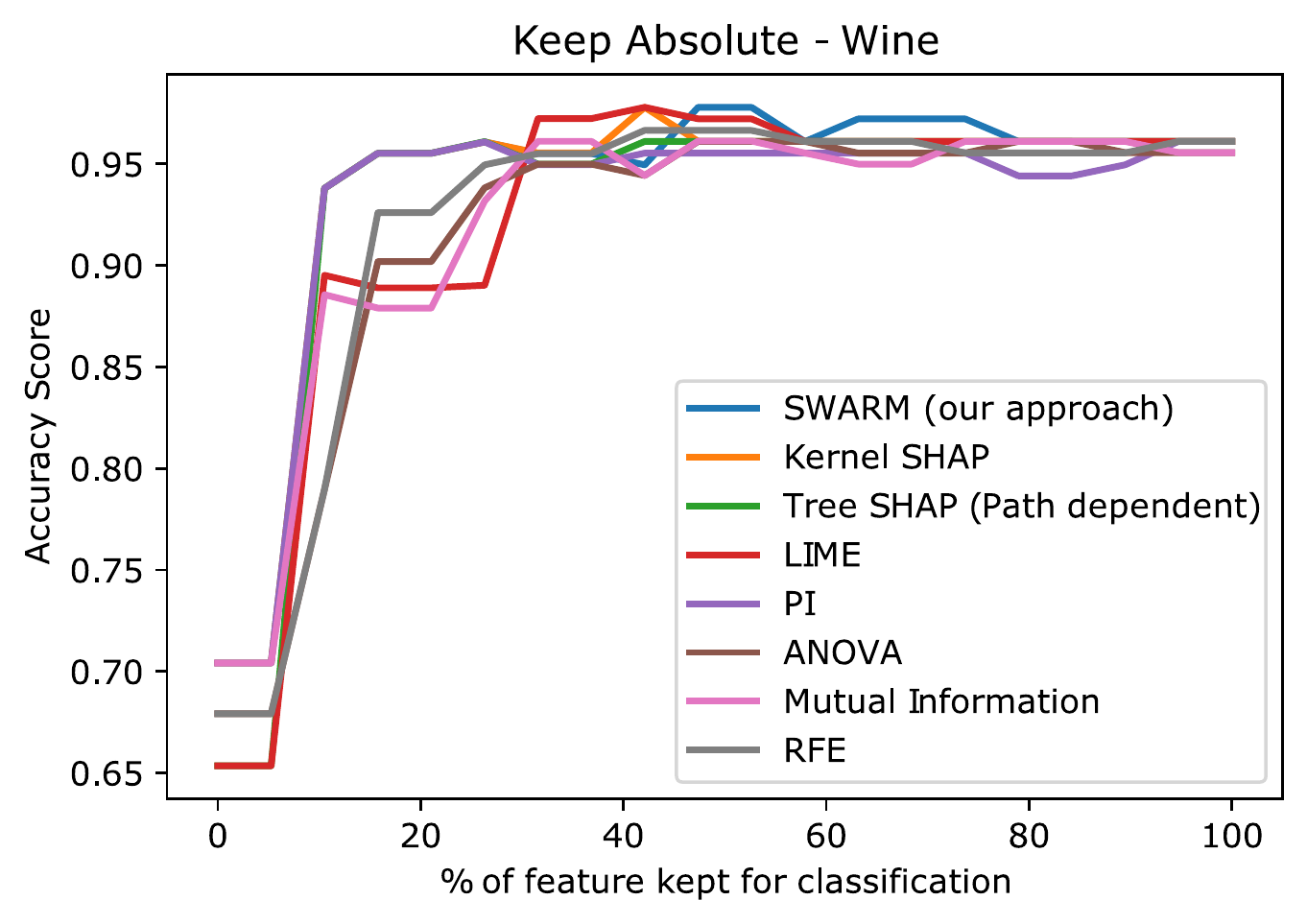}
 \caption{F1-Score when varying the number of features kept for classification.}
 \label{fig:keep-absolute}
\end{figure}

In this section, we compare our proposal against two well-established techniques at their ability to retrieve important features. The techniques were evaluated using the Keep Absolute~\citep{Lundberg2020} metric, which measures the impact of selected features on the model accuracy. In this case, the impact is measured by adding the most important features from the dataset and training the model. Figure~\ref{fig:keep-absolute} exemplifies the strategy for the Keep Absolute metric, note that as more features are added, the accuracy of the model on cross-validation (5 fold) setting increases. Also, to further validate our methodology, we evaluate it against feature selection algorithms (Permutation Importance~\citep{Altmann2010}, ANOVA~\citep{Pedregosa2016}, Mutual Information~\citep{Ross2014}, and Recursive Feature Selection~\citep{Guyon2002}) using the same metric.

Table~\ref{tab:keep_absolute-xgboost} shows the values for the Keep Absolute metric using the XGBoost classifier. Our technique performed better than all of the feature selection techniques---although for the Vertebral dataset presented a difference of only 0.0009 to the best result. Focusing only on the model explanation techniques, our method provides explanation with better ordering of feature importance than at least one state-of-the-art method. From these results, we emphasize two important things. Firstly, our method performs very similar to well-established techniques in the literature. Second, the small differences among scores in Table~\ref{tab:keep_absolute-xgboost} shows great agreements among the model explanation approaches.

%Table~\ref{tab:keep_absolute-xgboost} shows that our technique was able to give the best results only for the \textbf{Wine} and \textbf{Iris} datasets. The latter, however, is a relatively simple data set that do not require much robustness from the techniques, as seen in the results. Although our technique was not able to uncover the best ordering of features according to the \textbf{Keep Absolute} metric, it presented better results than all of the feature selection algorithms for most of the dataset (for \textbf{Vertebral} dataset, however, our technique presented a difference of only $0.0009$ to the best result). Focusing on the model explanation techniques, our method was able to provide explanations that resulted in better feature ordering based on importance better than at least one method for all of the datasets (of course, excluding \textbf{Iris} dataset). From this numerical analysis, we could see that our method performs very similar to well-established methods in the literature. Besides that, looking at the differences among the scores in Table~\ref{tab:keep_absolute-xgboost}, it is possible to note that there are only a few differences in the ordering imposed by the algorithms.

\begin{table*}[h!]
\centering
\caption{Area Under the Curve (AUC) for Keep Absolute metric. The curve is calculated by varying the number of features kept for classification---the best scores are highlighted in bold.}
\label{tab:keep_absolute-xgboost}
\begin{tabular}{lcccccc}
\toprule 
               & Vertebral       & Indian Liver    & Heart           & Wine            & Breast Cancer   & Iris            \\
\midrule 
SWARM          & 0.8071          & 0.6687          & 0.8172          & \textbf{0.9413} & 0.9639          & \textbf{0.9615} \\
Kernel SHAP    & 0.8047          & 0.6672          & 0.8108          & 0.9393          & 0.9628          & \textbf{0.9615} \\
Tree SHAP (PD) & 0.8056          & \textbf{0.6692} & 0.8194          & 0.9338          & \textbf{0.9650} & \textbf{0.9615} \\
LIME           & 0.8074          & 0.6658          & 0.8197          & 0.9253          & 0.9645          & \textbf{0.9615} \\
PI             & \textbf{0.8080} & 0.6636          & \textbf{0.8209} & 0.9333          & 0.9637          & \textbf{0.9615} \\
ANOVA          & 0.7849          & 0.6632          & 0.7940          & 0.9187          & 0.9510          & \textbf{0.9615} \\
M. Information & 0.7933          & 0.6603          & 0.7922          & 0.9239          & 0.9505          & \textbf{0.9615} \\
RFE            & 0.7960          & 0.6585          & 0.7795          & 0.9246          & 0.9617          & 0.9560          \\
\bottomrule 
\end{tabular}

\end{table*}

As seen for XGBoost Classifier in Table~\ref{tab:keep_absolute-xgboost}, we also evaluated the techniques by using Random Forest Classifier. Table~\ref{tab:keep_absolute-rf} summarizes the results. We can see from the table that although our technique was not able to present the best results for all of the datasets, it presented the second-best score for \textit{Indian Liver}, \textit{Heart}, and \textit{Breast Cancer}. Finally, the significative difference was only reported for the \textit{Wine} dataset.

\begin{table*}[]
\centering
\caption{Area Under the Curve (AUC) for Keep Absolute metric and Random Forest Classifier.}
\label{tab:keep_absolute-rf}
\begin{tabular}{lcccccc}
\toprule 
               & Vertebral & Indian Liver & Heart  & Wine   & Breast Cancer & Iris   \\ 
\midrule
SWARM          & \textbf{0.8155}    & 0.6756       & 0.8178 & 0.9208 & 0.9563        & 0.9541 \\ 
Kernel SHAP    & 0.8128    & \textbf{0.6820}       & 0.8165 & \textbf{0.9433} & 0.9537        & 0.9535 \\ 
LIME           & 0.7931    & 0.6723       & \textbf{0.8212} & 0.9332 & 0.9559        & 0.9551 \\ 
PI             & 0.8144    & 0.6666       & 0.8115 & 0.9283 & \textbf{0.9566}        & 0.9551 \\ 
ANOVA          & 0.7916    & 0.6681       & 0.7897 & 0.9282 & 0.9490        & 0.9534 \\ 
M. Information & 0.7978    & 0.6652       & 0.7906 & 0.9332 & 0.9474        & 0.9551 \\ 
RFE            & 0.7966    & 0.6706       & 0.7740 & 0.9407 & 0.9514        & \textbf{0.9602} \\ 
\bottomrule 
\end{tabular}

\end{table*}

To illustrate the stability of our method in returning stable results compared with other techniques, we show the rankings of each method according to XGBoost Classifier and Random Forest Classifier. Based on the results for XGBoost Classifier in Table~\ref{tab:random-forest-orderings}, our method has a mean ranking of $2.33$ (underlined) while losing only for TreeSHAP (in bold) with a mean ranking of $2.17$. Finally, the results in Table~\ref{tab:xgboost-orderings} for Random Forest Classifier show that our method got the third position with $0.17$ behind the first two methods (KernelSHAP and LIME, both with a mean ranking of $3.00$). These results confirm the stability of the proposed technique.

\begin{table*}[h!]
\centering
\caption{Ranking for the results presented in Table~\ref{tab:keep_absolute-xgboost}.}
\label{tab:xgboost-orderings}
\begin{tabular}{lcccccccc}
\toprule 
               & Vertebral       & Indian Liver    & Heart           & Wine            & Breast Cancer   & Iris & Mean ranking & St.d.             \\
\midrule 
SWARM          & 3 & 2 & 4 & 1 & 3 & 1 & \underline{2.33} &	1.21 \\
Kernel SHAP    & 5 & 3 & 5 & 2 & 5 & 1 & 3.50 &	1.76 \\
Tree SHAP (PD) & 4 & 1 & 3 & 3 & 1 & 1 & \textbf{2.17} &	1.33 \\
LIME           & 2 & 4 & 2 & 5 & 2 & 1 & 2.67 &	1.51 \\
PI             & 1 & 5 & 1 & 4 & 4 & 1 & 2.67 &	1.86 \\
ANOVA          & 8 & 6 & 6 & 8 & 7 & 1 & 6.00 &	2.61 \\
M. Information & 7 & 7 & 7 & 7 & 8 & 1 & 6.17 &	2.56 \\
RFE            & 6 & 8 & 8 & 6 & 6 & 8 & 7.00 &	1.10 \\
\bottomrule 
\end{tabular}

\end{table*}

\begin{table*}[h!]
\centering
\caption{Ranking for the results presented in Table~\ref{tab:keep_absolute-rf}.}
\label{tab:random-forest-orderings}
\begin{tabular}{lcccccccc}
\toprule 
               & Vertebral       & Indian Liver    & Heart           & Wine            & Breast Cancer   & Iris  & Mean ranking & St.d.           \\
\midrule 
SWARM & 1 &	2 &	2 &	7 &	2 &	5	&	\underline{3.17}	&	2.32 \\
Kernel SHAP & 3 &	1 &	3 &	1 &	4 &	6	&	\textbf{3.00}	&	1.90 \\
LIME & 6 &	3 &	1 &	3 &	3 &	2	&	\textbf{3.00}	&	1.67 \\
PI & 2 &	6 &	4 &	5 &	1 &	2	&	3.33	&	1.97 \\
ANOVA & 7 &	5 &	6 &	6 &	6 &	7	&	6.17	&	0.75 \\
M. Information & 4 &	7 &	5 &	3 &	7 &	2	&	4.67	&	2.07 \\
RFE & 5 &	4 &	7 &	2 &	5 &	1	&	4.00	&	2.19 \\

\bottomrule 
\end{tabular}

\end{table*}

\iffalse 
To illustrate how our method is stable in returning good results compared with other techniques, we use the distribution of the position of each technique for the datasets of Tables~\ref{tab:keep_absolute-xgboost} and~\ref{tab:keep_absolute-rf}. The distributions in Figure~\ref{fig:distribution-orderings} show that the methods designed for explainability present stability while presenting good results, i.e., they present results that assume positions from the first to the third positions. Regarding our technique, it presented very similar results to SHAP for both classifiers. Finally, it is worth noticing that our method was only the second-best for Random Forest Classifier (see Fig.~\ref{fig:distribution-orderings-rf}) since it provided the worst ordering for the \textit{Wine} dataset, however, it provided -- in general -- the best for the other datasets.

\begin{figure}[!htb]
    \centering
    \subfloat[Distribution of orderings using XGBoost Classifier.]{\includegraphics[width=0.5\columnwidth]{position_distribution_xgboost.pdf}\label{fig:distribution-orderings-xgboost}}    
    \subfloat[Distribution of orderings using Random Forest Classifier.]{\includegraphics[width=0.5\columnwidth]{position_distribution_rf.pdf}\label{fig:distribution-orderings-rf}}
    \caption{Position of the techniques after ordering based on AUC.}
    \label{fig:distribution-orderings}
\end{figure}
\fi

%\subsection{Sensitivity test}

%\lipsum[1-2]

\subsection{Parameter analysis}

\begin{figure*}[htb!]
 \centering % avoid the use of \begin{center}...\end{center} and use \centering instead (more compact)
 \includegraphics[width=0.8\linewidth]{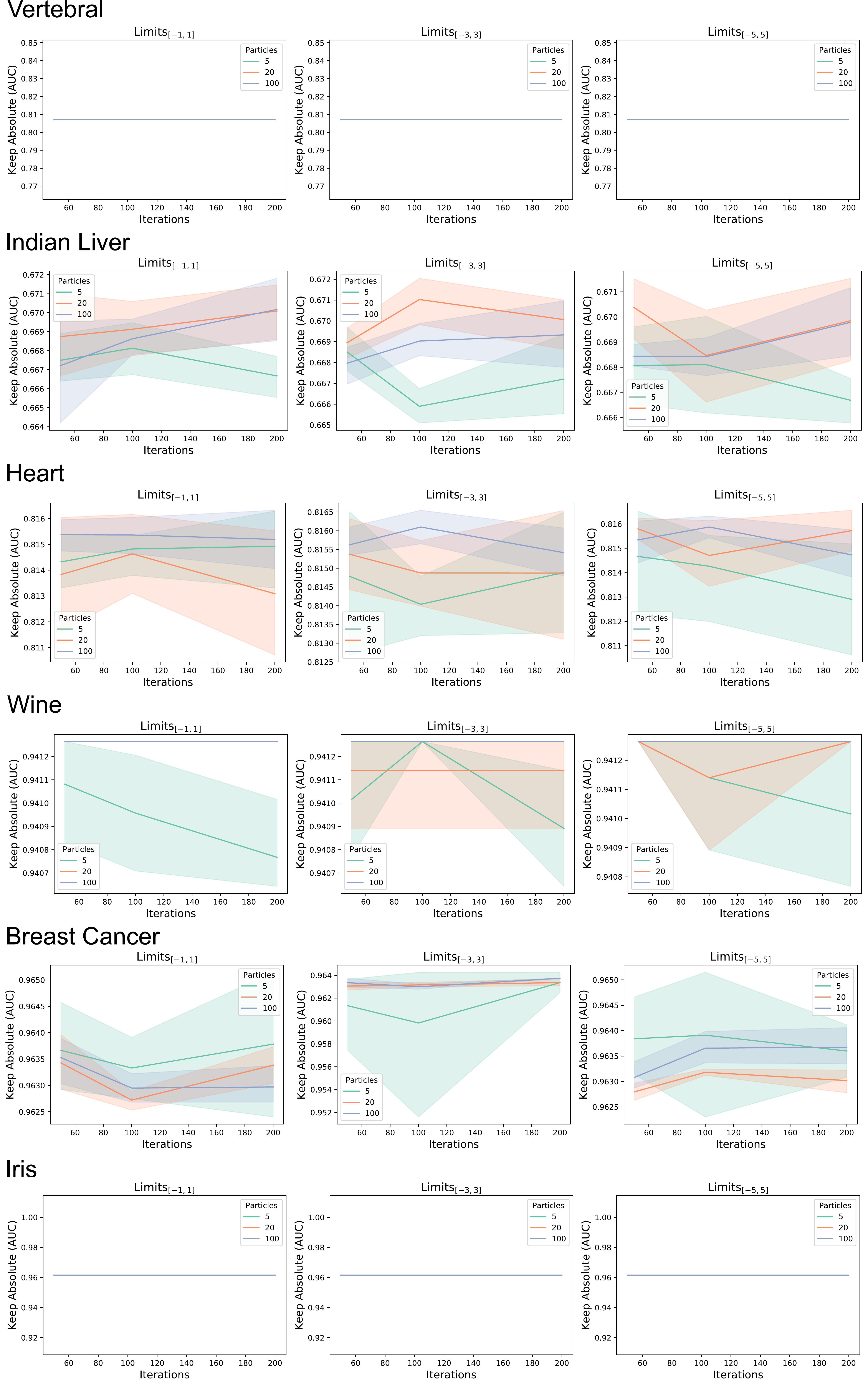}
 \caption{Keep Absolute metric (AUC) when varying the PSO parameters.}
 \label{fig:parameter_analysis}
\end{figure*}

In this section, we provide an evaluation on a few parameters of PSO algorithm and their influence on the result of our model. We evaluate the velocity parameters ($V_{\text{min}}$, $V_{\text{max}}$)---indicated by Limits$_{[V_{min}, V_{max}]}$ in the graphs---, which influence the changes of particles in each algorithm iteration. The number of particles ($p_n$) used by the swarm optimization, and the number of iterations. Figure~\ref{fig:parameter_analysis} shows the result for the datasets used in the numerical evaluation using the Keep Absolute metric after five runs using the same configuration.

The results of Figure~\ref{fig:parameter_analysis} show the robustness of our model regarding the changes in the parameters. That is, although there are differences among the Keep Absolute (AUC) values when considering different parameter combinations, the overall performance is maintained. Another important thing to mention is that the number of iterations together with the number of particles define the runtime execution for each dataset (please, see \textbf{Figure 1} in \textbf{Supplementary File}). Thus, the analysis of Figure~\ref{fig:parameter_analysis} provides the information that using a few particles (< 10) is sufficient for a trustful analysis using our model.

\section{Discussions}
\label{sec:discussion}

In this work, we presented a novel model interpretation approach using PSO. Our method, defined as a global interpretation approach, can be employed as a local approach by feeding one instance at a time to our PSO implementation.  The main strength of our method is its simplicity and ability to be adapted to various performance scores and optimization functions, which is a strength of our method over the state-of-the-art algorithms LIME~\citep{Ribeiro2016} and SHAP~\citep{Lundberg2017}.

In the numerical evaluation, our method performed better than feature selection algorithms on tasks that measure the performance of defining the input features' proper importance ordering. Although it could not uncover the best results for all of the datasets used in training, our method provided stable results on higher positions. 

\paragraph{\textbf{Preprocessing.}} One important thing to mention is the need to preprocess some datasets to use the proposed approach with PSO perturbation weights. For instance, a $\epsilon$ value needs to be added to datasets with zero entries. Such a preprocessing step is important since our explanations consist of using instructions in the form $w\times X_{ij}$, and having $X_{ij} = 0$ will make the particles of the PSO algorithm stay still and not make any progress. Adding $\epsilon$ to $X$ will solve such an problem---here we used $\epsilon = 1$. Notice that this preprocessing is only applied to help the algorithm, and the visualization aspects are created based on the original data. 

Finally, while we demonstrated our technique's ability to explain classifiers' predictions, it can be easily modified to work with regression models. In this case, the PSO algorithm can be used to produce perturbation weights that would maximize an error metric (e.g., mean squared error) while reducing the distance between the position with no perturbation ($w = 1$).

\paragraph{\textbf{Other nature-inspired algorithms.}} As demonstrated in the paper, the PSO and the visualization design help understand the learned patterns by machine learning techniques. However, the reader may notice that the optimization performed to search for the perturbing features cannot be performed only by PSO. Other algorithms, such as genetic or evolutionary approaches, could also be employed for this task. The two components of our approach (optimization and visualization) are uncoupled and provide flexibility on which algorithm to use. Thus, we plan to investigate other nature-inspired strategies with a few updates on the visualization component to explain machine learning techniques in future works.

\subsection{Limitations}

The main limitation of our work is the run-time execution since we have to execute the PSO algorithm for each combination of feature and class. Although we plan to investigate faster optimization algorithms in further works, a subset of the data used to feed our algorithm could also decrease the execution time. Besides that, we could use a subset of the most important features for a dataset with higher dimensionality.

Another limitation of our work is the need for preprocessing steps, as discussed above, which could break the desire patterns introduced by users or during dataset acquisition. In future works, we plan to investigate alternatives to reduce the dependency on these steps. For example, instead of multiplying a feature by an optimized value, each data sample could receive a random perturbation (negative or positive) created using a Gaussian function with an optimized kernel. So, less important features could receive a strong perturbation without a relevant change in the model's performance. On the other hand, an essential feature would cause performance loss even with a weak perturbation.

\subsection{Implementation}

Our technique was implemented using Python, while the prototype system with the visualization uses D3.js~\citep{Bostock2011}. We also created a Python module (it will be available after publication) with lightweight visualizations similar to those presented here so users can use our approach in notebooks.

\section{Conclusion}
\label{sec:conclusion}

As machine learning algorithms take over traditional approaches to solve problems, their reliability becomes more critical in applications where wrong decisions can lead to severe issues. Understanding a model's decisions is now an essential process of the development and execution of machine learning strategies to assess if such decisions make sense to domain experts.

In this work, we proposed a novel model-agnostic approach to interpret any classification algorithm by using PSO. We validated our approach on its ability to explain classifiers' decisions throughout several case studies. Finally, our methodology was also numerically evaluated on its ability to classify input features according to their importance. The results showed that our method could return very stable orderings with good results for all of the datasets---a result that only one state-of-the-art method was able to provide.

We plan to investigate our methodology on regression tasks in future works and use other optimization algorithms to find the perturbation weights. Besides that, since SHAP approximates Shapley values, we also want to investigate nature-inspired optimization techniques to approximate them.

\section*{Acknowledgements}

This work was supported by the agencies Fundação de Amparo à Pesquisa (FAPESP)---grant \#2018/17881-3, and Coordenação de Aperfeiçoamento de Pessoal de Nível Superior (CAPES)---grant \#88887.487331/2020-00.

%% Loading bibliography style file
\bibliographystyle{model1-num-names}
% \bibliographystyle{cas-model2-names}

% Loading bibliography database
\bibliography{cas-refs}

\begin{thebibliography}{51}
\expandafter\ifx\csname natexlab\endcsname\relax\def\natexlab#1{#1}\fi
\providecommand{\bibinfo}[2]{#2}
\ifx\xfnm\relax \def\xfnm[#1]{\unskip,\space#1}\fi
%Type = Inproceedings
\bibitem[{Peters et~al.(2018)Peters, Neumann, Iyyer, Gardner, Clark, Lee, and
  Zettlemoyer}]{Peters2018}
\bibinfo{author}{M.~E. Peters}, \bibinfo{author}{M.~Neumann},
  \bibinfo{author}{M.~Iyyer}, \bibinfo{author}{M.~Gardner},
  \bibinfo{author}{C.~Clark}, \bibinfo{author}{K.~Lee},
  \bibinfo{author}{L.~Zettlemoyer},
\newblock \bibinfo{title}{Deep contextualized word representations},
\newblock in: \bibinfo{booktitle}{Proc. of NAACL}.
%Type = Misc
\bibitem[{Devlin et~al.(2018)Devlin, Chang, Lee, and Toutanova}]{Devlin2018}
\bibinfo{author}{J.~Devlin}, \bibinfo{author}{M.-W. Chang},
  \bibinfo{author}{K.~Lee}, \bibinfo{author}{K.~Toutanova},
  \bibinfo{title}{Bert: Pre-training of deep bidirectional transformers for
  language understanding}, \bibinfo{year}{2018}.
%Type = Inproceedings
\bibitem[{{Szegedy} et~al.(2015){Szegedy}, {Wei Liu}, {Yangqing Jia},
  {Sermanet}, {Reed}, {Anguelov}, {Erhan}, {Vanhoucke}, and
  {Rabinovich}}]{Szegedy2015}
\bibinfo{author}{C.~{Szegedy}}, \bibinfo{author}{{Wei Liu}},
  \bibinfo{author}{{Yangqing Jia}}, \bibinfo{author}{P.~{Sermanet}},
  \bibinfo{author}{S.~{Reed}}, \bibinfo{author}{D.~{Anguelov}},
  \bibinfo{author}{D.~{Erhan}}, \bibinfo{author}{V.~{Vanhoucke}},
  \bibinfo{author}{A.~{Rabinovich}},
\newblock \bibinfo{title}{Going deeper with convolutions},
\newblock in: \bibinfo{booktitle}{2015 IEEE Conference on Computer Vision and
  Pattern Recognition (CVPR)}, pp. \bibinfo{pages}{1--9}.
%Type = Incollection
\bibitem[{Krizhevsky et~al.(2012)Krizhevsky, Sutskever, and
  Hinton}]{Krizhevsky2016}
\bibinfo{author}{A.~Krizhevsky}, \bibinfo{author}{I.~Sutskever},
  \bibinfo{author}{G.~E. Hinton},
\newblock \bibinfo{title}{Imagenet classification with deep convolutional
  neural networks},
\newblock in: \bibinfo{editor}{F.~Pereira}, \bibinfo{editor}{C.~J.~C. Burges},
  \bibinfo{editor}{L.~Bottou}, \bibinfo{editor}{K.~Q. Weinberger} (Eds.),
  \bibinfo{booktitle}{Advances in Neural Information Processing Systems 25},
  \bibinfo{publisher}{Curran Associates, Inc.}, \bibinfo{year}{2012}, pp.
  \bibinfo{pages}{1097--1105}.
%Type = Inproceedings
\bibitem[{{He} et~al.(2016){He}, {Zhang}, {Ren}, and {Sun}}]{He2016}
\bibinfo{author}{K.~{He}}, \bibinfo{author}{X.~{Zhang}},
  \bibinfo{author}{S.~{Ren}}, \bibinfo{author}{J.~{Sun}},
\newblock \bibinfo{title}{Deep residual learning for image recognition},
\newblock in: \bibinfo{booktitle}{2016 IEEE Conference on Computer Vision and
  Pattern Recognition (CVPR)}, pp. \bibinfo{pages}{770--778}.
%Type = Inproceedings
\bibitem[{{Chollet}(2017)}]{Chollet2017}
\bibinfo{author}{F.~{Chollet}},
\newblock \bibinfo{title}{Xception: Deep learning with depthwise separable
  convolutions},
\newblock in: \bibinfo{booktitle}{2017 IEEE Conference on Computer Vision and
  Pattern Recognition (CVPR)}, pp. \bibinfo{pages}{1800--1807}.
%Type = Article
\bibitem[{Luo(2016)}]{Luo2016}
\bibinfo{author}{G.~Luo},
\newblock \bibinfo{title}{Automatically explaining machine learning prediction
  results: a demonstration on type 2 diabetes risk prediction},
\newblock \bibinfo{journal}{Health Information Science and Systems}
  \bibinfo{volume}{4} (\bibinfo{year}{2016}).
%Type = Inproceedings
\bibitem[{Bansal et~al.(2019)Bansal, Nushi, Kamar, Weld, Lasecki, and
  Horvitz}]{Bansal2019}
\bibinfo{author}{G.~Bansal}, \bibinfo{author}{B.~Nushi},
  \bibinfo{author}{E.~Kamar}, \bibinfo{author}{D.~S. Weld},
  \bibinfo{author}{W.~S. Lasecki}, \bibinfo{author}{E.~Horvitz},
\newblock \bibinfo{title}{Updates in human-ai teams: Understanding and
  addressing the performance/compatibility tradeoff},
\newblock in: \bibinfo{booktitle}{AAAI}.
%Type = Inproceedings
\bibitem[{Kulesza et~al.(2015)Kulesza, Burnett, Wong, and Stumpf}]{Kulesza2015}
\bibinfo{author}{T.~Kulesza}, \bibinfo{author}{M.~Burnett},
  \bibinfo{author}{W.-K. Wong}, \bibinfo{author}{S.~Stumpf},
\newblock \bibinfo{title}{Principles of explanatory debugging to personalize
  interactive machine learning},
\newblock in: \bibinfo{booktitle}{Proceedings of the 20th International
  Conference on Intelligent User Interfaces}, IUI ’15,
  \bibinfo{publisher}{ACM}, \bibinfo{year}{2015}, p.
  \bibinfo{pages}{126–137}.
%Type = Article
\bibitem[{Doshi-Velez and Kim(2017)}]{DoshiVelez2017}
\bibinfo{author}{F.~Doshi-Velez}, \bibinfo{author}{B.~Kim},
\newblock \bibinfo{title}{Towards a rigorous science of interpretable machine
  learning},
\newblock \bibinfo{journal}{arXiv}  (\bibinfo{year}{2017}).
%Type = Inproceedings
\bibitem[{Doshi-Velez and Kim(2018)}]{DoshiVelez2018}
\bibinfo{author}{F.~Doshi-Velez}, \bibinfo{author}{B.~Kim},
\newblock \bibinfo{title}{Considerations for evaluation and generalization in
  interpretable machine learning}.
%Type = Inproceedings
\bibitem[{Krause et~al.(2016)Krause, Perer, and Ng}]{Krause2016}
\bibinfo{author}{J.~Krause}, \bibinfo{author}{A.~Perer},
  \bibinfo{author}{K.~Ng},
\newblock \bibinfo{title}{Interacting with predictions: Visual inspection of
  black-box machine learning models},
\newblock in: \bibinfo{booktitle}{Proceedings of the 2016 CHI Conference on
  Human Factors in Computing Systems}, CHI ’16, \bibinfo{publisher}{ACM},
  \bibinfo{year}{2016}, p. \bibinfo{pages}{5686–5697}.
%Type = Article
\bibitem[{{Pezzotti} et~al.(2018){Pezzotti}, {Höllt}, {Van Gemert},
  {Lelieveldt}, {Eisemann}, and {Vilanova}}]{Pezzotti2018}
\bibinfo{author}{N.~{Pezzotti}}, \bibinfo{author}{T.~{Höllt}},
  \bibinfo{author}{J.~{Van Gemert}}, \bibinfo{author}{B.~P.~F. {Lelieveldt}},
  \bibinfo{author}{E.~{Eisemann}}, \bibinfo{author}{A.~{Vilanova}},
\newblock \bibinfo{title}{Deepeyes: Progressive visual analytics for designing
  deep neural networks},
\newblock \bibinfo{journal}{IEEE Transactions on Visualization and Computer
  Graphics} \bibinfo{volume}{24} (\bibinfo{year}{2018})
  \bibinfo{pages}{98--108}.
%Type = Inproceedings
\bibitem[{Marcilio-Jr et~al.(2020)Marcilio-Jr, Eler, Garcia, Correia, and
  Silva}]{MarcilioJr2020}
\bibinfo{author}{W.~E. Marcilio-Jr}, \bibinfo{author}{D.~M. Eler},
  \bibinfo{author}{R.~E. Garcia}, \bibinfo{author}{R.~C.~M. Correia},
  \bibinfo{author}{L.~F. Silva},
\newblock \bibinfo{title}{A hybrid visualization approach to perform analysis
  of feature spaces},
\newblock in: \bibinfo{editor}{S.~Latifi} (Ed.), \bibinfo{booktitle}{17th
  International Conference on Information Technology--New Generations (ITNG
  2020)}, \bibinfo{publisher}{Springer International Publishing},
  \bibinfo{address}{Cham}, \bibinfo{year}{2020}, pp. \bibinfo{pages}{241--247}.
%Type = Incollection
\bibitem[{Lundberg and Lee(2017)}]{Lundberg2017}
\bibinfo{author}{S.~M. Lundberg}, \bibinfo{author}{S.-I. Lee},
\newblock \bibinfo{title}{A unified approach to interpreting model
  predictions},
\newblock in: \bibinfo{editor}{I.~Guyon}, \bibinfo{editor}{U.~V. Luxburg},
  \bibinfo{editor}{S.~Bengio}, \bibinfo{editor}{H.~Wallach},
  \bibinfo{editor}{R.~Fergus}, \bibinfo{editor}{S.~Vishwanathan},
  \bibinfo{editor}{R.~Garnett} (Eds.), \bibinfo{booktitle}{Advances in Neural
  Information Processing Systems 30}, \bibinfo{publisher}{Curran Associates,
  Inc.}, \bibinfo{year}{2017}, pp. \bibinfo{pages}{4765--4774}.
%Type = Inproceedings
\bibitem[{Ribeiro et~al.(2016)Ribeiro, Singh, and Guestrin}]{Ribeiro2016}
\bibinfo{author}{M.~T. Ribeiro}, \bibinfo{author}{S.~Singh},
  \bibinfo{author}{C.~Guestrin},
\newblock \bibinfo{title}{"why should {I} trust you?": Explaining the
  predictions of any classifier},
\newblock in: \bibinfo{booktitle}{Proceedings of the 22nd {ACM} {SIGKDD}
  International Conference on Knowledge Discovery and Data Mining, San
  Francisco, CA, USA, August 13-17, 2016}, pp. \bibinfo{pages}{1135--1144}.
%Type = Article
\bibitem[{{Zhang} et~al.(2019){Zhang}, {Wang}, {Molino}, {Li}, and
  {Ebert}}]{Zhang2019}
\bibinfo{author}{J.~{Zhang}}, \bibinfo{author}{Y.~{Wang}},
  \bibinfo{author}{P.~{Molino}}, \bibinfo{author}{L.~{Li}},
  \bibinfo{author}{D.~S. {Ebert}},
\newblock \bibinfo{title}{Manifold: A model-agnostic framework for
  interpretation and diagnosis of machine learning models},
\newblock \bibinfo{journal}{IEEE Transactions on Visualization and Computer
  Graphics} \bibinfo{volume}{25} (\bibinfo{year}{2019})
  \bibinfo{pages}{364--373}.
%Type = Article
\bibitem[{{Hinterreiter} et~al.(2020){Hinterreiter}, {Ruch}, {Stitz},
  {Ennemoser}, {Bernard}, {Strobelt}, and {Streit}}]{Hinterreiter2020}
\bibinfo{author}{A.~{Hinterreiter}}, \bibinfo{author}{P.~{Ruch}},
  \bibinfo{author}{H.~{Stitz}}, \bibinfo{author}{M.~{Ennemoser}},
  \bibinfo{author}{J.~{Bernard}}, \bibinfo{author}{H.~{Strobelt}},
  \bibinfo{author}{M.~{Streit}},
\newblock \bibinfo{title}{Confusionflow: A model-agnostic visualization for
  temporal analysis of classifier confusion},
\newblock \bibinfo{journal}{IEEE Transactions on Visualization and Computer
  Graphics}  (\bibinfo{year}{2020}) \bibinfo{pages}{1--1}.
%Type = Book
\bibitem[{Olsson(2010)}]{Olsson2010}
\bibinfo{author}{A.~E. Olsson}, \bibinfo{title}{Particle Swarm Optimization:
  Theory, Techniques and Applications}, \bibinfo{publisher}{Nova Science
  Publishers, Inc.}, \bibinfo{address}{USA}, \bibinfo{year}{2010}.
%Type = Article
\bibitem[{Balagopalan et~al.(2018)Balagopalan, Novikova, Rudzicz, and
  Ghassemi}]{Balagopalan2018}
\bibinfo{author}{A.~Balagopalan}, \bibinfo{author}{J.~Novikova},
  \bibinfo{author}{F.~Rudzicz}, \bibinfo{author}{M.~Ghassemi},
\newblock \bibinfo{title}{The effect of heterogeneous data for alzheimer's
  disease detection from speech},
\newblock \bibinfo{journal}{ArXiv} \bibinfo{volume}{abs/1811.12254}
  (\bibinfo{year}{2018}).
%Type = Article
\bibitem[{Esteva et~al.(2017)Esteva, Kuprel, Novoa, Ko, Swetter, Blau, and
  Thrun}]{Esteva2017}
\bibinfo{author}{A.~Esteva}, \bibinfo{author}{B.~Kuprel},
  \bibinfo{author}{R.~A. Novoa}, \bibinfo{author}{J.~Ko},
  \bibinfo{author}{S.~M. Swetter}, \bibinfo{author}{H.~M. Blau},
  \bibinfo{author}{S.~Thrun},
\newblock \bibinfo{title}{Dermatologist-level classification of skin cancer
  with deep neural networks},
\newblock \bibinfo{journal}{Nature} \bibinfo{volume}{542}
  (\bibinfo{year}{2017}) \bibinfo{pages}{115--118}.
%Type = Article
\bibitem[{Modarres et~al.(2018)Modarres, Ibrahim, Louie, and
  Paisley}]{Modarres2018}
\bibinfo{author}{C.~Modarres}, \bibinfo{author}{M.~Ibrahim},
  \bibinfo{author}{M.~Louie}, \bibinfo{author}{J.~W. Paisley},
\newblock \bibinfo{title}{Towards explainable deep learning for credit lending:
  A case study},
\newblock \bibinfo{journal}{ArXiv} \bibinfo{volume}{abs/1811.06471}
  (\bibinfo{year}{2018}).
%Type = Article
\bibitem[{Meijer and Wessels(2019)}]{Meijer2019}
\bibinfo{author}{A.~Meijer}, \bibinfo{author}{M.~Wessels},
\newblock \bibinfo{title}{Predictive policing: Review of benefits and
  drawbacks},
\newblock \bibinfo{journal}{International Journal of Public Administration}
  \bibinfo{volume}{42} (\bibinfo{year}{2019}) \bibinfo{pages}{1031--1039}.
%Type = Article
\bibitem[{Hong et~al.(2020)Hong, Hullman, and Bertini}]{Hong2020}
\bibinfo{author}{S.~R. Hong}, \bibinfo{author}{J.~Hullman},
  \bibinfo{author}{E.~Bertini},
\newblock \bibinfo{title}{Human factors in model interpretability: Industry
  practices, challenges, and needs},
\newblock \bibinfo{journal}{Proc. ACM Hum.-Comput. Interact.}
  \bibinfo{volume}{4} (\bibinfo{year}{2020}).
%Type = Article
\bibitem[{Lundberg et~al.(2020)Lundberg, Erion, Chen, DeGrave, Prutkin, Nair,
  Katz, Himmelfarb, Bansal, and Lee}]{Lundberg2020}
\bibinfo{author}{S.~M. Lundberg}, \bibinfo{author}{G.~Erion},
  \bibinfo{author}{H.~Chen}, \bibinfo{author}{A.~DeGrave},
  \bibinfo{author}{J.~M. Prutkin}, \bibinfo{author}{B.~Nair},
  \bibinfo{author}{R.~Katz}, \bibinfo{author}{J.~Himmelfarb},
  \bibinfo{author}{N.~Bansal}, \bibinfo{author}{S.-I. Lee},
\newblock \bibinfo{title}{From local explanations to global understanding with
  explainable ai for trees},
\newblock \bibinfo{journal}{Nature Machine Intelligence} \bibinfo{volume}{2}
  (\bibinfo{year}{2020}) \bibinfo{pages}{2522--5839}.
%Type = Article
\bibitem[{Baehrens et~al.(2010)Baehrens, Schroeter, Harmeling, Kawanabe,
  Hansen, and M\"{u}ller}]{Baehrens2010}
\bibinfo{author}{D.~Baehrens}, \bibinfo{author}{T.~Schroeter},
  \bibinfo{author}{S.~Harmeling}, \bibinfo{author}{M.~Kawanabe},
  \bibinfo{author}{K.~Hansen}, \bibinfo{author}{K.-R. M\"{u}ller},
\newblock \bibinfo{title}{How to explain individual classification decisions},
\newblock \bibinfo{journal}{J. Mach. Learn. Res.} \bibinfo{volume}{11}
  (\bibinfo{year}{2010}) \bibinfo{pages}{1803–1831}.
%Type = Article
\bibitem[{Strumbelj and Kononenko(2013)}]{Strumbelj2013}
\bibinfo{author}{E.~Strumbelj}, \bibinfo{author}{I.~Kononenko},
\newblock \bibinfo{title}{Explaining prediction models and individual
  predictions with feature contributions},
\newblock \bibinfo{journal}{Knowledge and Information Systems}
  \bibinfo{volume}{41} (\bibinfo{year}{2013}) \bibinfo{pages}{647--665}.
%Type = Inproceedings
\bibitem[{{Datta} et~al.(2016){Datta}, {Sen}, and {Zick}}]{Datta2016}
\bibinfo{author}{A.~{Datta}}, \bibinfo{author}{S.~{Sen}},
  \bibinfo{author}{Y.~{Zick}},
\newblock \bibinfo{title}{Algorithmic transparency via quantitative input
  influence: Theory and experiments with learning systems},
\newblock in: \bibinfo{booktitle}{2016 IEEE Symposium on Security and Privacy
  (SP)}, pp. \bibinfo{pages}{598--617}.
%Type = Article
\bibitem[{Smilkov et~al.(2017)Smilkov, Carter, Sculley, Vi{\'e}gas, and
  Wattenberg}]{Smilkov2017}
\bibinfo{author}{D.~Smilkov}, \bibinfo{author}{S.~Carter},
  \bibinfo{author}{D.~Sculley}, \bibinfo{author}{F.~B. Vi{\'e}gas},
  \bibinfo{author}{M.~Wattenberg},
\newblock \bibinfo{title}{Direct-manipulation visualization of deep networks},
\newblock \bibinfo{journal}{ArXiv} \bibinfo{volume}{abs/1708.03788}
  (\bibinfo{year}{2017}).
%Type = Article
\bibitem[{Kahng et~al.(2018)Kahng, Andrews, Kalro, and Polo~Chau}]{Kahng2018}
\bibinfo{author}{M.~Kahng}, \bibinfo{author}{P.~Y. Andrews},
  \bibinfo{author}{A.~Kalro}, \bibinfo{author}{D.~H. Polo~Chau},
\newblock \bibinfo{title}{Activis: Visual exploration of industry-scale deep
  neural network models},
\newblock \bibinfo{journal}{IEEE transactions on visualization and computer
  graphics} \bibinfo{volume}{24} (\bibinfo{year}{2018})
  \bibinfo{pages}{88—97}.
%Type = Article
\bibitem[{{Rauber} et~al.(2017){Rauber}, {Fadel}, {Falcão}, and
  {Telea}}]{Rauber2017}
\bibinfo{author}{P.~E. {Rauber}}, \bibinfo{author}{S.~G. {Fadel}},
  \bibinfo{author}{A.~X. {Falcão}}, \bibinfo{author}{A.~C. {Telea}},
\newblock \bibinfo{title}{Visualizing the hidden activity of artificial neural
  networks},
\newblock \bibinfo{journal}{IEEE Transactions on Visualization and Computer
  Graphics} \bibinfo{volume}{23} (\bibinfo{year}{2017})
  \bibinfo{pages}{101--110}.
%Type = Inproceedings
\bibitem[{{Clavien} et~al.(2019){Clavien}, {Alberti}, {Pondenkandath},
  {Ingold}, and {Liwicki}}]{Clavien2019}
\bibinfo{author}{G.~{Clavien}}, \bibinfo{author}{M.~{Alberti}},
  \bibinfo{author}{V.~{Pondenkandath}}, \bibinfo{author}{R.~{Ingold}},
  \bibinfo{author}{M.~{Liwicki}},
\newblock \bibinfo{title}{Dnnviz: Training evolution visualization for deep
  neural network},
\newblock in: \bibinfo{booktitle}{2019 6th Swiss Conference on Data Science
  (SDS)}, pp. \bibinfo{pages}{19--24}.
%Type = Article
\bibitem[{{Lex} et~al.(2014){Lex}, {Gehlenborg}, {Strobelt}, {Vuillemot}, and
  {Pfister}}]{Lex2014}
\bibinfo{author}{A.~{Lex}}, \bibinfo{author}{N.~{Gehlenborg}},
  \bibinfo{author}{H.~{Strobelt}}, \bibinfo{author}{R.~{Vuillemot}},
  \bibinfo{author}{H.~{Pfister}},
\newblock \bibinfo{title}{Upset: Visualization of intersecting sets},
\newblock \bibinfo{journal}{IEEE Transactions on Visualization and Computer
  Graphics} \bibinfo{volume}{20} (\bibinfo{year}{2014})
  \bibinfo{pages}{1983--1992}.
%Type = Inproceedings
\bibitem[{Chen and Guestrin(2016)}]{Chen2016}
\bibinfo{author}{T.~Chen}, \bibinfo{author}{C.~Guestrin},
\newblock \bibinfo{title}{Xgboost: A scalable tree boosting system},
\newblock KDD '16, \bibinfo{publisher}{Association for Computing Machinery},
  \bibinfo{address}{New York, NY, USA}, \bibinfo{year}{2016}, p.
  \bibinfo{pages}{785–794}.
%Type = Misc
\bibitem[{Dua and Graff(2017)}]{Dua2019}
\bibinfo{author}{D.~Dua}, \bibinfo{author}{C.~Graff}, \bibinfo{title}{{UCI}
  machine learning repository}, \bibinfo{year}{2017}.
%Type = Article
\bibitem[{Labelle et~al.(2005)Labelle, Roussouly, Berthonnaud, Dimnet, and
  Obrien}]{Labelle2005}
\bibinfo{author}{H.~Labelle}, \bibinfo{author}{P.~Roussouly},
  \bibinfo{author}{E.~Berthonnaud}, \bibinfo{author}{J.~Dimnet},
  \bibinfo{author}{M.~Obrien},
\newblock \bibinfo{title}{The importance of spino-pelvic balance in l5–s1
  developmental spondylolisthesis: A review of pertinent radiologic
  measurements},
\newblock \bibinfo{journal}{Spine} \bibinfo{volume}{30} (\bibinfo{year}{2005})
  \bibinfo{pages}{S27--S34}.
%Type = Article
\bibitem[{Roussouly and Pinheiro-Franco(2011{\natexlab{a}})}]{Roussouly2011}
\bibinfo{author}{P.~Roussouly}, \bibinfo{author}{J.~L. Pinheiro-Franco},
\newblock \bibinfo{title}{Biomechanical analysis of the spino-pelvic
  organization and adaptation in pathology},
\newblock \bibinfo{journal}{European Spine Journal} \bibinfo{volume}{20}
  (\bibinfo{year}{2011}{\natexlab{a}}) \bibinfo{pages}{609--618}.
%Type = Article
\bibitem[{Roussouly and Pinheiro-Franco(2011{\natexlab{b}})}]{Huec2011}
\bibinfo{author}{P.~Roussouly}, \bibinfo{author}{J.~L. Pinheiro-Franco},
\newblock \bibinfo{title}{Sagittal spino-pelvic balance is a crucial analysis
  for normal and degenerative spine},
\newblock \bibinfo{journal}{Eur Spine J} \bibinfo{volume}{20}
  (\bibinfo{year}{2011}{\natexlab{b}}) \bibinfo{pages}{556–557}.
%Type = Article
\bibitem[{Breiman(2001)}]{Breiman2001}
\bibinfo{author}{L.~Breiman},
\newblock \bibinfo{title}{Random forests},
\newblock \bibinfo{journal}{Machine Learning} \bibinfo{volume}{45}
  (\bibinfo{year}{2001}) \bibinfo{pages}{5--32}.
%Type = Misc
\bibitem[{GJameson~JL(2019)}]{Jameson2019}
\bibinfo{author}{e.~a. GJameson~JL}, \bibinfo{title}{Disorders of hemoglobin},
  \bibinfo{howpublished}{Harrison's Principles of Internal Medicine.},
  \bibinfo{year}{2019}.
%Type = Misc
\bibitem[{Tha(2019)}]{Thalassemias2019}
\bibinfo{title}{National heart, lung, and blood institute},
  \bibinfo{howpublished}{Harrison's Principles of Internal Medicine.},
  \bibinfo{year}{2019}.
%Type = Book
\bibitem[{Gersh(2000)}]{Gersh2000}
\bibinfo{author}{B.~J. Gersh}, \bibinfo{title}{Mayo Clinic Heart Book},
  \bibinfo{publisher}{HarperCollins}, \bibinfo{edition}{2\textsuperscript{nd}}
  edition, \bibinfo{year}{2000}.
%Type = Article
\bibitem[{Khatibi and Montazer(2010)}]{Khatibi2010}
\bibinfo{author}{V.~Khatibi}, \bibinfo{author}{G.~A. Montazer},
\newblock \bibinfo{title}{A fuzzy-evidential hybrid inference engine for
  coronary heart disease risk assessment},
\newblock \bibinfo{journal}{Expert Systems with Applications}
  \bibinfo{volume}{37} (\bibinfo{year}{2010}) \bibinfo{pages}{8536 -- 8542}.
%Type = Article
\bibitem[{Dorogush et~al.(2018)Dorogush, Ershov, and Gulin}]{Dorogush2018}
\bibinfo{author}{A.~V. Dorogush}, \bibinfo{author}{V.~Ershov},
  \bibinfo{author}{A.~Gulin},
\newblock \bibinfo{title}{Catboost: gradient boosting with categorical features
  support},
\newblock \bibinfo{journal}{ArXiv} \bibinfo{volume}{abs/1810.11363}
  (\bibinfo{year}{2018}).
%Type = Article
\bibitem[{Abdul-Ghani and DeFronzo(2009)}]{AbdulGhani2010}
\bibinfo{author}{M.~A. Abdul-Ghani}, \bibinfo{author}{R.~A. DeFronzo},
\newblock \bibinfo{title}{Plasma glucose concentration and prediction of future
  risk of type 2 diabetes},
\newblock \bibinfo{journal}{Diabetes care} \bibinfo{volume}{32}
  (\bibinfo{year}{2009}) \bibinfo{pages}{194--198}.
%Type = Article
\bibitem[{Bays et~al.(2007)Bays, Chapman, and Grandy}]{Bays2007}
\bibinfo{author}{H.~E. Bays}, \bibinfo{author}{R.~H. Chapman},
  \bibinfo{author}{S.~Grandy},
\newblock \bibinfo{title}{The relationship of body mass index to diabetes
  mellitus, hypertension and dyslipidaemia: comparison of data from two
  national surveys},
\newblock \bibinfo{journal}{International journal of clinical practice}
  \bibinfo{volume}{61} (\bibinfo{year}{2007}) \bibinfo{pages}{737--747}.
%Type = Article
\bibitem[{Altmann et~al.(2010)Altmann, Tolosi, Sander, and
  Lengauer}]{Altmann2010}
\bibinfo{author}{A.~Altmann}, \bibinfo{author}{L.~Tolosi},
  \bibinfo{author}{O.~Sander}, \bibinfo{author}{T.~Lengauer},
\newblock \bibinfo{title}{Permutation importance: a corrected feature
  importance measure.},
\newblock \bibinfo{journal}{Bioinform.} \bibinfo{volume}{26}
  (\bibinfo{year}{2010}) \bibinfo{pages}{1340--1347}.
%Type = Article
\bibitem[{Pedregosa et~al.(2011)Pedregosa, Varoquaux, Gramfort, Michel,
  Thirion, Grisel, Blondel, Prettenhofer, Weiss, Dubourg
  et~al.}]{Pedregosa2016}
\bibinfo{author}{F.~Pedregosa}, \bibinfo{author}{G.~Varoquaux},
  \bibinfo{author}{A.~Gramfort}, \bibinfo{author}{V.~Michel},
  \bibinfo{author}{B.~Thirion}, \bibinfo{author}{O.~Grisel},
  \bibinfo{author}{M.~Blondel}, \bibinfo{author}{P.~Prettenhofer},
  \bibinfo{author}{R.~Weiss}, \bibinfo{author}{V.~Dubourg}, et~al.,
\newblock \bibinfo{title}{Scikit-learn: Machine learning in python},
\newblock \bibinfo{journal}{Journal of machine learning research}
  \bibinfo{volume}{12} (\bibinfo{year}{2011}) \bibinfo{pages}{2825--2830}.
%Type = Article
\bibitem[{Ross(2014)}]{Ross2014}
\bibinfo{author}{B.~C. Ross},
\newblock \bibinfo{title}{Mutual information between discrete and continuous
  data sets} \bibinfo{volume}{9} (\bibinfo{year}{2014})
  \bibinfo{pages}{e87357}.
%Type = Inproceedings
\bibitem[{Guyon et~al.(????)Guyon, Weston, Barnhill, Vapnik, and
  Cristianini}]{Guyon2002}
\bibinfo{author}{I.~Guyon}, \bibinfo{author}{J.~Weston},
  \bibinfo{author}{S.~Barnhill}, \bibinfo{author}{V.~Vapnik},
  \bibinfo{author}{N.~Cristianini},
\newblock \bibinfo{title}{Gene selection for cancer classification using
  support vector machines},
\newblock in: \bibinfo{booktitle}{Machine Learning}, p. \bibinfo{pages}{2002}.
%Type = Article
\bibitem[{{Bostock} et~al.(2011){Bostock}, {Ogievetsky}, and
  {Heer}}]{Bostock2011}
\bibinfo{author}{M.~{Bostock}}, \bibinfo{author}{V.~{Ogievetsky}},
  \bibinfo{author}{J.~{Heer}},
\newblock \bibinfo{title}{D3 data-driven documents},
\newblock \bibinfo{journal}{IEEE Transactions on Visualization and Computer
  Graphics} \bibinfo{volume}{17} (\bibinfo{year}{2011})
  \bibinfo{pages}{2301--2309}.

\end{thebibliography}

% \bio{figs/pic1} BLIND
% \endbio

\end{document}